\PassOptionsToPackage{table}{xcolor}
\PassOptionsToPackage{hyphens}{url}
\documentclass[10pt]{article}
\ifdefined\pdfcompresslevel\pdfcompresslevel=9\fi
\ifdefined\pdfobjcompresslevel\pdfobjcompresslevel=3\fi
\usepackage{renaissance-report}
\usepackage{algorithm}
\usepackage{algorithmic}
\usepackage{newfloat}
\usepackage{listings}
\usepackage{mdframed}
\DeclareCaptionStyle{ruled}{labelfont=normalfont,labelsep=colon,strut=off}
\floatstyle{ruled}
\newfloat{listing}{tb}{lst}{}
\floatname{listing}{Listing}
\usepackage{multirow}
\usepackage{longtable}
\usepackage{pifont}

\definecolor{DrawAIHeatBlue}{HTML}{4F81A6}
\definecolor{DrawAIHeatRed}{HTML}{B86666}
\definecolor{DrawAIHeatAmber}{HTML}{C58A3A}
\definecolor{DrawAIHeatGray}{HTML}{C9D0D5}
\definecolor{DrawAIPromptBackground}{HTML}{F3F6F7}
\definecolor{DrawAIPromptBorder}{HTML}{8EA3B0}

\lstdefinestyle{drawaipromptcode}{basicstyle={\scriptsize\ttfamily},
    backgroundcolor=\color{white},
    frame=single,
    rulecolor=\color{DrawAIPromptBorder},
    framerule=0.4pt,
    framesep=5pt,
    framexleftmargin=3pt,
    framexrightmargin=3pt,
    xleftmargin=0pt,
    xrightmargin=0pt,
    numbers=none,
    showstringspaces=false,
    columns=fullflexible,
    keepspaces=true,
    breaklines=true,
    breakatwhitespace=true,
    tabsize=2,
    aboveskip=5pt,
    belowskip=5pt}

\newmdenv[
    backgroundcolor=DrawAIPromptBackground,
    linecolor=DrawAIPromptBorder,
    linewidth=0.6pt,
    roundcorner=2pt,
    skipabove=8pt,
    skipbelow=8pt,
    innerleftmargin=10pt,
    innerrightmargin=10pt,
    innertopmargin=8pt,
    innerbottommargin=8pt,
    splittopskip=8pt,
    splitbottomskip=8pt
]{drawaipromptbox}

\newcommand{\drawaiprompttitle}[1]{\par\noindent{\large\bfseries #1}\par\vspace{0.35em}}
\newcommand{\drawaipromptheading}[1]{\par\vspace{0.55em}\noindent{\normalsize\bfseries #1}\par\vspace{0.15em}}
\providecommand{\tightlist}{\setlength{\itemsep}{2pt}\setlength{\parskip}{0pt}}

\DeclareRobustCommand{\modelopenai}[1]{\mbox{\raisebox{-0.12em}{\includegraphics[height=0.82em]{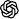}}\kern0.18em#1}}
\DeclareRobustCommand{\modelclaude}[1]{\mbox{\raisebox{-0.12em}{\includegraphics[height=0.82em]{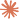}}\kern0.18em#1}}
\DeclareRobustCommand{\modelgemini}[1]{\mbox{\raisebox{-0.12em}{\includegraphics[height=0.82em]{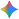}}\kern0.18em#1}}
\DeclareRobustCommand{\modelqwen}[1]{\mbox{\raisebox{-0.12em}{\includegraphics[height=0.82em]{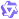}}\kern0.18em#1}}
\DeclareRobustCommand{\modelminimax}[1]{\mbox{\raisebox{-0.12em}{\includegraphics[height=0.82em]{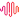}}\kern0.18em#1}}
\DeclareRobustCommand{\modelkimi}[1]{\mbox{\raisebox{-0.12em}{\includegraphics[height=0.82em]{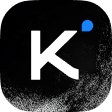}}\kern0.18em#1}}
\DeclareRobustCommand{\modelmimo}[1]{\mbox{\raisebox{-0.12em}{\includegraphics[height=0.82em]{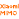}}\kern0.18em#1}}

\reporttype{Research Paper}
\title{DrawAI: Agentic Benchmark and Workflow for Making Raster Images Editable}
\author[1]{Pu Cao}
\author[1]{Qingye Kong}
\author[2]{Xuedan Yin}
\author[1]{Xuekun Zhao}
\author[1]{Rupeng Yan}
\author[1]{Qing Song}
\author[3]{Yao Zhang}
\author[1,\correspondingmark]{Lu Yang}
\affil[1]{Beijing University of Posts and Telecommunications}
\affil[2]{Tsinghua University}
\affil[3]{Beijing Institute of Technology}
\authornote{\correspondingmark\ Corresponding author}
\projectpage{https://drawai.renaissancemind.ai/}
\dataseturl{https://huggingface.co/datasets/caopu/DrawAI-Bench}
\codeurl{https://github.com/Renaissance-Mind/DrawAI}
\reportmeta{Contact}{\href{mailto:caopu@bupt.edu.cn,soeaver@bupt.edu.cn}{\texttt{\{caopu, soeaver\}@bupt.edu.cn}}}

\begin{document}

\maketitle

\begin{abstract}
Recent image-generation models and multimodal agents can produce high-quality visuals for increasingly complex visual communication tasks. Yet their raster outputs remain difficult to use directly because meaningful content and relationships are flattened into pixels, preventing users from inspecting, modifying, rearranging, or reusing individual components. We formulate \emph{image-to-editable reconstruction}, which recovers a structured, directly manipulable artifact from a raster image while preserving its visual and semantic content. The central challenge is to jointly satisfy \emph{Fidelity} and \emph{Editability}, which often trade off in practice. To study this task, we introduce DrawAI, comprising an agentic benchmark, \textbf{DrawAI-Bench}, and a reconstruction workflow, \textbf{DrawAI-Flow}. DrawAI-Bench spans scientific figures, presentation slides, posters, and diagrams, combining real and AI-generated images to reflect practical visual-creation scenarios. It evaluates Fidelity and Editability through a hybrid protocol of 39 criteria: deterministic rule-based metrics measure properties with direct correspondences, while asset-specific vision-language rubrics capture semantic and perceptual qualities for which exact matching is misleading. Besides, we propose DrawAI-Flow, a two-stage agentic workflow in which a Parser Agent turns extracted elements evidence into an explicit reconstruction plan, and a Reconstruction Agent realizes the plan as executable graphics code through an iterative code--render--validate--revise loop. On DrawAI-Bench, we systematically evaluate thirteen models across five agent harnesses to study the effects of model capability, harness choice, and workflow design. The results show that reconstruction quality and costs vary substantially across model--harness configurations, while DrawAI-Flow consistently improves editable structure.
\end{abstract}

\section{Introduction}
\begin{figure*}[t]
\centering
\includegraphics[width=\linewidth]{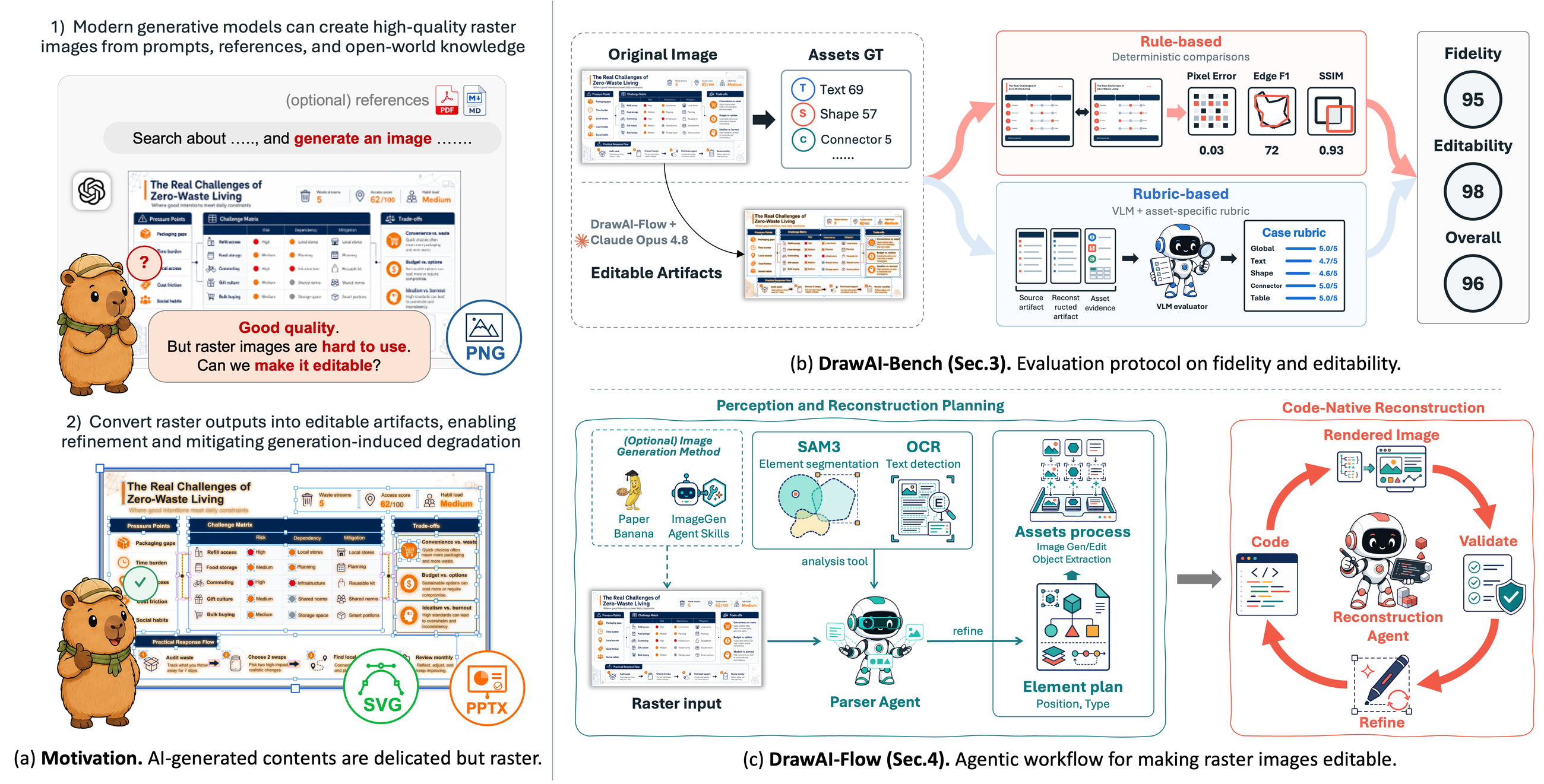}
\caption{\textbf{DrawAI at a glance.} (a) The image-to-editable reconstruction task recovers structured, directly manipulable SVG or PPTX artifacts from high-quality raster images. (b) DrawAI-Bench jointly evaluates visual Fidelity and structural Editability through a hybrid protocol of rule-based metrics and VLM rubrics. (c) DrawAI-Flow is a two-stage agentic workflow that translates perception evidence into an explicit reconstruction plan and realizes it through iterative image-as-code synthesis.}
\label{fig:teaser}
\end{figure*}

Recent image-generation models have moved visual synthesis beyond open-ended image creation toward practical content production, producing high-quality visuals and structured compositions from natural-language instructions \citep{ramesh2021zeroshot,nichol2021glide,saharia2022imagen,rombach2022latent}. Coupled with agents, this capability now extends to complex visual communication tasks. PaperBanana and AutoFigure can produce publication-ready scientific illustrations \citep{zhu2026paperbanana,zhu2026autofigure}, while NotebookLM turns collections of sources into a range of visual artifacts, including slide decks, infographics, and mind maps. Such outputs are increasingly useful not only as final images, but also as working drafts that users expect to correct, rearrange, and reuse. Yet these generated outputs remain difficult to use directly, as illustrated in Figure~\ref{fig:teaser}(a). Raster outputs collapse content, object structure, and spatial relationships into a single pixel canvas, whereas meaningful revisions target individual elements and their relations. Even a local correction can therefore require manual redrawing or full regeneration, risking changes to regions that users have already accepted. It is therefore critical to transform visually compelling generated content into human-friendly editable artifacts whose meaningful components can be directly inspected, modified, rearranged, and reused.

In this work, we formulate and systematically investigate \emph{image-to-editable reconstruction}. Given a high-quality raster image, the task is to recover a structured artifact whose deterministic rendering preserves the visual and semantic content of the source while exposing its constituent elements for direct manipulation. The central difficulty is that visual resemblance and editability are closely related but often trade off in practice. The source image can be reproduced almost perfectly by embedding it as a full-page bitmap, but doing so exposes no usable structure. Conversely, decomposing the image into many editable objects can omit content, distort layout, break semantic relations, or reduce a distinctive visual design to generic primitives. Recent advances in multimodal agents make this challenge increasingly approachable \citep{yao2023react,yang2023mmreact,wang2024openhands}. Agents can inspect images, invoke perception tools, generate executable graphics code, and revise rendered outputs through visual feedback. These capabilities turn image-to-editable reconstruction from a fixed conversion problem into an agentic process whose success criteria, execution strategy, and capability requirements can now be studied systematically.

Motivated by this opportunity, we introduce \textbf{DrawAI} and investigate the following questions: 1) how can we construct an evaluation framework that measures Fidelity and Editability jointly? 2) how can a purpose-built agentic workflow empower multimodal agents to produce stronger reconstructions? 3) how well do current agents perform, and how do model choice, execution harness, and workflow design affect their results? Together, these questions connect evaluation design, agent empowerment, and the empirical diagnosis of current systems.

To construct this evaluation framework, we develop \textbf{DrawAI-Bench} over scientific figures, presentation slides, posters, and diagrams, as shown in Figure~\ref{fig:teaser}(b). These domains cover complementary editing and composition demands, including information-dense explanation, mixed text--graphic layout, expressive visual composition, and explicit relational structure. We include both real and AI-generated images. Real images reflect established human design conventions, whereas AI-generated images directly instantiate the generation-to-editable conversion process that motivates this work, enabling us to test the feasibility of an end-to-end automated visual-authoring pipeline. To combine reproducibility with semantic and perceptual coverage, DrawAI-Bench introduces a hybrid evaluation protocol of 39 criteria over Fidelity and Editability. Its rule-based component measures properties with direct correspondences that admit deterministic computation, while its rubric-based component uses asset-specific vision-language model judgments where exact matching would be misleading. Together, the two components provide a more comprehensive assessment of how reconstructed artifacts perform in practice.

To execute the task and study how structured support affects agents, we further develop \textbf{DrawAI-Flow}, the two-stage process shown in Figure~\ref{fig:teaser}(c). A Parser Agent consolidates SAM3 and OCR evidence, corrects the initial decomposition, and records element-specific decisions in a structured reconstruction plan. A Reconstruction Agent then realizes that plan as executable graphics code and iterates through rendering, validation, and revision. Separating interpretation from iterative realization reduces the burden on a single end-to-end generation step and makes intermediate decisions inspectable and correctable. Rather than directly emitting the complete target artifact, DrawAI-Flow uses executable graphics code as an intermediate representation of it. This choice aligns reconstruction with agents' code-generation and iterative-repair capabilities, while deterministic execution materializes the editable artifact for rendering and validation \citep{wang2024codeact,rodriguez2025starvector}. Because revisions can be expressed as localized code edits, this representation can reduce low-level format errors and avoid repeatedly spending tokens on unchanged structure.

On DrawAI-Bench, we systematically evaluate thirteen models across five agent harnesses to study the effects of model capability, harness choice, and workflow design. The results show that reconstruction quality varies substantially across model--harness configurations, while DrawAI-Flow consistently improves editable structure. Reliable image-to-editable reconstruction is therefore shaped by the complete model--harness--workflow system rather than by the model alone.

Our contributions are summarized as follows:
\begin{itemize}
    \item We introduce DrawAI-Bench, a cross-domain benchmark that jointly evaluates visual Fidelity and practical Editability through a hybrid protocol of 39 rule-based and rubric-based criteria.
    \item We develop DrawAI-Flow, a two-stage agentic workflow that turns multimodal perception evidence into an explicit reconstruction plan and realizes it through iterative, validated image-as-code reconstruction.
    \item We conduct a controlled study across models, harnesses, reconstruction settings, specialized projects, and costs, revealing that robust reconstruction depends on the complete execution system and remains substantially aided by structured workflow support.
\end{itemize}

\section{Related Work}

\subsection{Controllable Image Generation}

\paragraph{Text-to-image generation} has progressed from large-scale autoregressive models to diffusion- and flow-based systems with strong prompt following, photorealism, typography, and instruction following \citep{ramesh2021zeroshot,nichol2021glide,saharia2022imagen,rombach2022latent,esser2024rectified,wu2025qwenimage}. Controllable generation further conditions these models on boxes, sketches, edges, depth, poses, reference images, and adapters \citep{li2023gligen,zhang2023controlnet,mou2024t2iadapter,ye2023ipadapter,cao2025controllable,cao2025imageallneed}. 
While these control mechanisms offer precise guidance over the generation process, the final outputs remain fundamentally constrained as flattened raster images.

\paragraph{Agentic visual content generation} further improves information representation. FigGen distinguishes text-to-figure generation from natural-image synthesis \citep{rodriguez2023figgen}; PaperBanana and AutoFigure use agentic planning and refinement to generate publication-ready figures and evaluate faithfulness, readability, aesthetics, and completeness \citep{zhu2026paperbanana,zhu2026autofigure}. AutoFigure-Edit, Crafter, and VisPainter move toward native SVG editing and structured visual composition \citep{lin2026autofigureedit,zhao2026crafter,sun2025pixels}. A parallel line automates presentations and posters: DOC2PPT and PPTAgent generate slide decks from documents \citep{fu2022doc2ppt,zheng2025pptagent}, Paper2Poster produces editable PPTX posters from papers \citep{pang2025paper2poster}, and dedicated benchmarks score generated decks on content, aesthetics, and editability \citep{yang2026slidesgenbench,chen2026presentbench,jang2026deckbench}. DrawAI further aims to recover editable structures directly from existing images, introducing a cross-domain benchmark that explicitly couples visual fidelity with object-level editability.

\subsection{Image-to-Editable Artifacts}

A long line of work translates images into structured outputs for one format at a time: markup for formulas \citep{deng2017markup}, plotting code for charts \citep{yang2025chartmimic}, HTML for interfaces \citep{beltramelli2017pix2code,si2025design2code,yun2024web2code}, and paths or SVG programs via vectorization and vision-language SVG generation \citep{li2020diffvg,carlier2020deepsvg,ma2022live,hu2024supersvg,rodriguez2025starvector,yang2025omnisvg,nishina2024svgeditbench}. Recent and concurrent efforts recover editable structure from existing rasters within single domains: SliDer and Images2Slides derender slides and infographics into semantic SVG or native slide objects \citep{hazimeh2025slider,gonzalez2026images2slides}, AmodalSVG and AnchorFlow vectorize natural images into independently editable layers and sparsely anchored paths \citep{hu2026amodalsvg,jiang2026anchorflow}, and emerging benchmarks assess diagram-as-code and layered design authoring \citep{su2026vcgbench,deganutti2026gdb,jeong2025canvas}. 
Moving beyond format-specific vectorization, DrawAI formulates a systematic, cross-domain evaluation that jointly scores visual fidelity and object-level editability, thereby diagnosing the true capacity of modern agents to produce practically reusable artifacts.

\section{DrawAI-Bench}
\label{sec:benchmark}

\begin{figure*}[!t]
\centering
\includegraphics[width=\textwidth]{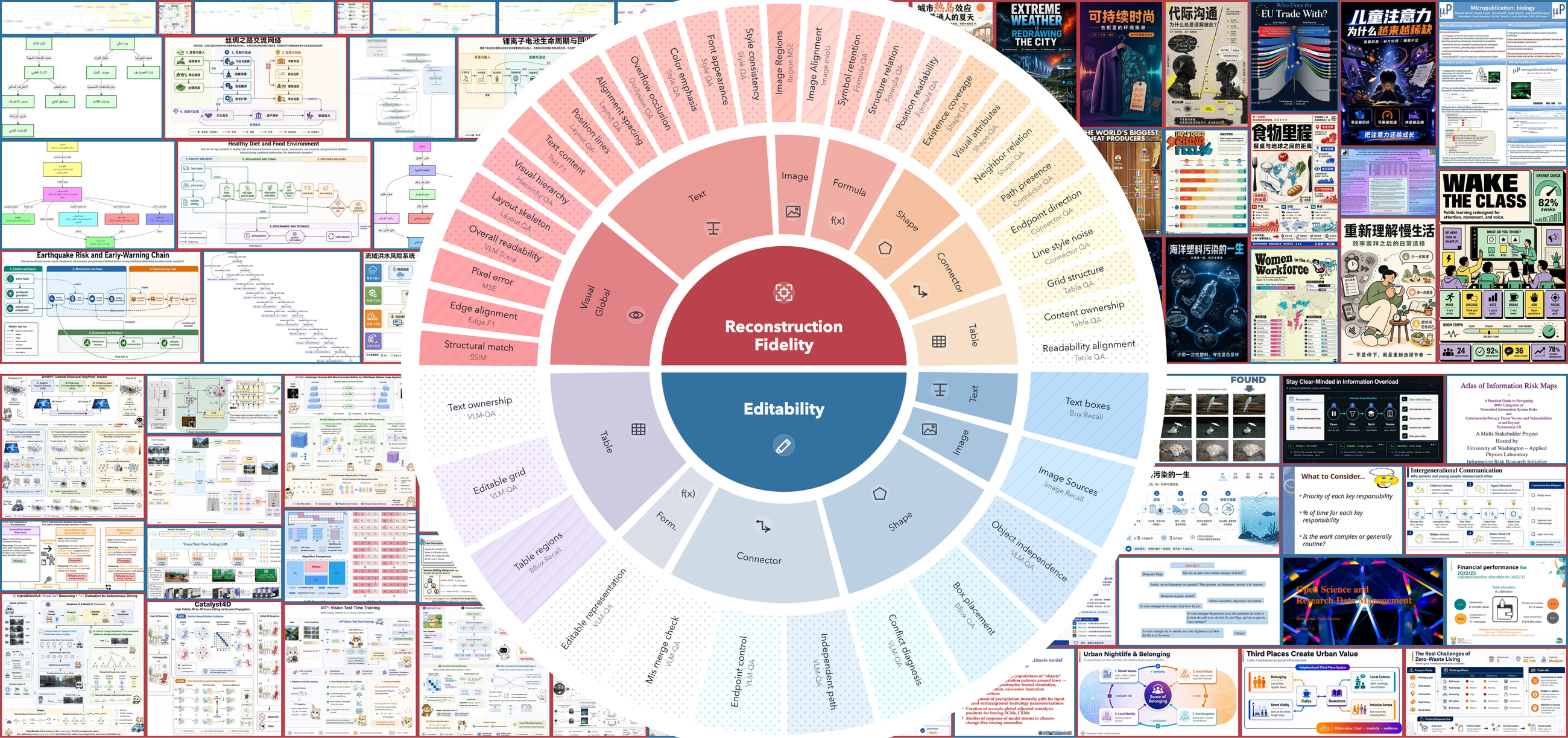}
\caption{\textbf{DrawAI-Bench data and evaluation taxonomy.} Examples from the four domains surround the metric hierarchy. Fidelity evaluates global and asset-specific reconstruction, while Editability tests whether text, images, formulas, shapes, connectors, and tables remain structured authoring objects. In the outer ring, diagonal-hatched backgrounds denote rule-based metrics, while dotted backgrounds denote rubric-based judgments. Images outlined in red and blue represent AI-generated and real examples, respectively.}
\label{fig:eval-taxonomy}
\end{figure*}

DrawAI-Bench aims to rigorously measure the effectiveness of image-to-editable reconstruction in real-world use cases. It covers four prevalent visual-content categories and includes both real-world and AI-generated images, allowing the benchmark to faithfully assess performance under realistic production conditions. The benchmark composition and evaluation taxonomy are illustrated in Figure~\ref{fig:eval-taxonomy}.

\subsection{Data Overview}

We collect images from four domains: scientific figures, presentation slides, posters, and diagrams. Together, these domains capture complementary authoring demands, ranging from information-dense explanation and mixed text--graphic layouts to expressive visual composition and explicit relational structure. 
They therefore represent common settings in which reconstructed artifacts are expected to support correction and reuse, rather than viewing alone. 
Each domain contains 10 real images and 10 AI-generated images, balancing mature human design conventions with outputs that directly instantiate the generation-to-editable transition. Scientific figures are sampled from published papers and outputs of figure-generation systems \citep{zhu2026paperbanana,zhu2026autofigure,lin2026autofigureedit,zhao2026crafter}. Presentation slides and posters are drawn from public slide corpora and collections of scientific and information posters, respectively, and paired with newly generated counterparts. For these generated counterparts, an agent first gathers content across diverse topics, after which image-generation models synthesize the corresponding visuals using style prompts sampled from an online gallery. Finally, the diagram domain includes flowcharts, mind maps, and structured explanatory graphics collected from public resources or produced by generation models.

We manually screen images for legibility, content diversity, and meaningful editable structure. To enable element-level evaluation, we annotate the asset type and spatial extent of every element according to the taxonomy introduced in Section~\ref{sec:evaluation-protocol}. GPT-5.5 first produces initial type and localization annotations following a dedicated annotation guideline. We then manually inspect and correct them using a purpose-built annotation refinement system. We denote each benchmark item as $(I_i,G_i)$, where $I_i$ is the source image and $G_i$ its verified element annotations. Beyond documenting the image content, $G_i$ determines which asset-specific criteria apply and provides element-level references for the evaluation described next.

\subsection{Evaluation Protocol}
\label{sec:evaluation-protocol}

Given $(I_i,G_i)$, a system synthesizes an editable artifact $A_i$, which can be deterministically rendered as $R_i=\textsc{Render}(A_i)$. DrawAI-Bench compares the rendered $R_i$ with the source $I_i$ for visual fidelity, while directly inspecting the structural topology of $A_i$ with $G_i$ to assess editability. This joint optimization guards against two degenerate shortcuts: embedding the source as a monolithic raster block to spoof visual similarity, or producing hyper-fragmented editable objects that fail to preserve the original visual semantics. Accordingly, the protocol first defines what should be measured through two complementary dimensions, and then specifies how each criterion is evaluated using rule- and rubric-based evidence.

\paragraph{Evaluation Dimensions}

Traditional evaluations often optimize for either visual resemblance or object decomposition. DrawAI-Bench explicitly measures the tension between them through two complementary dimensions: \emph{Fidelity} and \emph{Editability}, structured into 39 tertiary criteria as illustrated in Figure~\ref{fig:eval-taxonomy}. Across both dimensions, the taxonomy covers six recurring asset types: text, images, formulas, shapes, connectors, and tables.

Fidelity evaluates the rendered result at both the global and asset levels. The Global category captures whole-composition properties, including structural match, edge alignment, readability, and visual hierarchy. Asset-specific categories assess whether each type preserves the content, appearance, placement, and relationships appropriate to it. Conversely, Editability evaluates the underlying architecture of the artifact: whether the reconstructed content remains semantically grouped, cleanly separable, and directly manipulable. Its criteria examine properties like native text-box preservation, independent connector paths, controllable endpoints, and explicit table-grid structures. These categories provide comprehensive coverage of core authoring structures, ensuring the evaluated artifacts reflect practical downstream utility.

\paragraph{Hybrid Evaluation}
To systematically score these 39 criteria, DrawAI-Bench employs a hybrid framework that aligns the evaluation method with each property's underlying semantics. Properties with explicit geometric or textual correspondences are measured deterministically, whereas semantic organization and authoring intent require contextual judgment. DrawAI-Bench therefore combines 9 rule-based metrics with 30 rubric-based judgments. Both evaluation branches draw dynamically from the evaluation tuple $(I_i, R_i, G_i, A_i)$, selecting exactly the evidence required for the specific criterion.

The rule-based component computes objective distances and overlaps, utilizing metrics such as SSIM, pixel and region MSE, edge F1, OCR-normalized text F1, image mIoU, and bounding-box recall. In contrast, the rubric-based component leverages a vision-language model (VLM) to assess nuanced properties where strict exact-matching would penalize valid functional reconstructions. This includes evaluating visual hierarchy, formula relations, connector directionality, semantic grouping, and raster crop localization. The VLM assessments are mapped to a standardized discrete scale (\emph{yes}$=1$, \emph{partial}$=0.5$, and \emph{no}$=0$).

After direction normalization to $[0,1]$, we hierarchically average valid scores from applicable elements to criteria, categories, and dimensions, and compute Overall as the mean of Fidelity and Editability.

\section{DrawAI-Flow}
\label{sec:flow}

\subsection{Design Overview}

Image-to-editable reconstruction couples two distinct challenges: inferring an authoring structure from flattened pixels, and synthesizing that structure into a valid, manipulable artifact. Forcing a model to solve both in a single end-to-end generation step conflates perception errors with low-level formatting failures. DrawAI-Flow resolves this by decomposing the task into two specialized, agent-driven stages (Figure~\ref{fig:teaser}c). 

\subsection{Parsing and Planning}

The parsing stage must translate uncertain perceptual outputs into explicit authoring directives. To establish a foundation, we employ robust vision foundation models, specifically SAM3~\citep{carion2025sam3} for region proposal and PP-OCRv5~\citep{cui2025paddleocr} for text recognition, to generate initial candidate elements. Because these raw proposals are frequently incomplete, duplicated, or semantically fragmented, a \emph{Parser Agent} is then introduced to refine the structured reconstruction plan according to the preliminary evidence.

The agent reconciles this evidence against the holistic source image through spatial deduplication, omission recovery, and semantic alignment (e.g., correcting misclassified regions). The finalized reconstruction plan records each element's identity, geometry, appearance, and intended reconstruction policy. This policy explicitly anchors the Fidelity--Editability tradeoff to concrete implementation choices: text, vectors, and formulas are rebuilt using native primitives, while complex textures or photographs use localized crops. Foreground assets that require isolation are processed with BRIA RMBG-2.0, whereas assets requiring content-aware repair or redrawing are edited with GPT-Image-2.

\subsection{Image-as-Code Reconstruction}

The reconstruction stage must translate the parser's plan into a highly faithful artifact without compromising its modifiable structure. Rather than forcing the agent to serialize a complex SVG or PPTX in a single pass, the \emph{Reconstruction Agent} formulates an executable graphics program and refines it iteratively.
This \emph{image-as-code} paradigm serves as a powerful intermediate representation. It shifts the synthesis task into a code-action space where agents can naturally compose modular functions, execute intermediate states, and reliably debug syntax \citep{wang2024codeact}. Deterministic execution maps the program to a semantic SVG, where an element registry binds planned identifiers directly to native SVG routines or approved local assets. 

Because the executable program persists across iterations, the system supports a code--\allowbreak render--\allowbreak validate--\allowbreak revise loop. In each of the maximum five rounds, DrawAI-Flow renders the current SVG, compares it against the source image, and diagnoses critical discrepancies across structural, semantic, and stylistic hierarchies. The validator then prompts the agent for targeted code revisions. This localized repair mechanism significantly reduces redundant token consumption and minimizes the risk of introducing cascading formatting errors. Finally, the approved semantic SVG is verified for compatibility and can be seamlessly exported to an editable format like PPTX.

\section{Experiments and Analysis}
\label{sec:experiments}

\subsection{Experimental Settings}
\label{sec:settings}

\paragraph{Models.}
We evaluate thirteen multimodal large-language model variants from seven providers: \modelopenai{\emph{OpenAI}:} GPT-5.6 Sol, GPT-5.5, and GPT-5.4; \modelclaude{\emph{Anthropic}:} Claude Fable 5 and Claude Opus 4.8; \modelgemini{\emph{Google}:} Gemini 3.5 Flash and Gemini 3.1 Pro; \modelqwen{\emph{Alibaba}:} Qwen3.7 Plus; \modelminimax{\emph{MiniMax}:} MiniMax M3; \modelkimi{\emph{Moonshot AI}:} Kimi K3 and Kimi K2.7 Code; and \modelmimo{\emph{Xiaomi}:} MiMo V2.5 Pro and MiMo V2.5.

\paragraph{Agent Harnesses.}
We run the models through five agent harnesses: Codex, OpenHands, Claude Code, Kimi Code, and Antigravity. The detailed model--harness coverage is given below.

\paragraph{Reconstruction Frameworks.}
\emph{DrawAI-Flow} uses the two-stage workflow in Section~\ref{sec:flow}. To probe native model--harness capability, we also include a \emph{Simple Baseline} that gives each agent only the source image and the shared reconstruction instruction in Appendix~\ref{sec:reconstruction-prompt}. We further compare DrawAI with AutoFigure-Edit~\citep{lin2026autofigureedit}, Crafter~\citep{zhao2026crafter}, and Edit-Banana.

\paragraph{Evaluation.}
Every completed output is evaluated with the DrawAI-Bench protocol in Section~\ref{sec:benchmark}. To apply its hybrid evaluation consistently across systems, we compute the rule-based component deterministically from the rendered artifact and editable file, while judging the rubric-based component independently with GPT-5.6 Sol Ultra.

\paragraph{Models and access.}
We evaluate thirteen provider-hosted model variants: \modelopenai{GPT-5.6 Sol}, \modelopenai{GPT-5.5}, and \modelopenai{GPT-5.4}; \modelclaude{Claude Fable 5} and \modelclaude{Claude Opus 4.8}; \modelgemini{Gemini 3.5 Flash} and \modelgemini{Gemini 3.1 Pro}; \modelqwen{Qwen3.7 Plus}; \modelminimax{MiniMax M3}; \modelkimi{Kimi K3} and \modelkimi{Kimi K2.7 Code}; and \modelmimo{MiMo V2.5 Pro} and \modelmimo{MiMo V2.5}. The Gemini variants use the \emph{High} profile exposed by their execution environment.

\paragraph{Model--harness coverage.}
Codex is evaluated with the three GPT variants. Claude Code is evaluated with the two Claude variants, Kimi K2.7 Code, MiniMax M3, Qwen3.7 Plus, and the two MiMo variants. Kimi Code is evaluated with Kimi K3 and Kimi K2.7 Code, and Antigravity with the two Gemini variants. OpenHands provides the common cross-provider harness for GPT-5.5, Kimi K2.7 Code, MiniMax M3, Qwen3.7 Plus, and the two MiMo variants. This overlap supports within-model harness comparisons without conflating harness effects with model identity.

\paragraph{Auxiliary systems.}
During benchmark construction, style prompts for AI-generated counterparts are sampled from an online visual-style gallery.\footnote{\url{https://visualize.renaissancemind.ai/styles}} DrawAI-Flow uses BRIA RMBG-2.0 for foreground isolation\footnote{\url{https://huggingface.co/briaai/RMBG-2.0}} and GPT-Image-2 for content-aware repair or redrawing.\footnote{\url{https://developers.openai.com/api/docs/models/gpt-image-2}}

\paragraph{Execution and evaluation conditions.}
Every candidate is produced in an isolated run directory with a one-hour wall-time limit. Within a condition, agents receive the same source raster, output contract, and benchmark split, and may write and execute local reconstruction code. Model-level results select the strongest observed harness for each model, whereas harness analyses retain only within-model comparisons over matched source images. All candidate SVGs are rendered through the same deterministic pipeline and evaluated with the 39-criterion protocol. Rule-based criteria use the rendering and editable SVG, while rubric-based criteria are judged independently by \modelopenai{GPT-5.6 Sol Ultra} with the questions specified in Appendix~\ref{sec:complete-evaluation-criteria}. Overall is the equal mean of Fidelity and Editability.

\subsection{Human Validation of DrawAI-Bench}
\label{sec:human-validation}

We conduct two human studies on the same balanced sample of 200 reconstructed artifacts. The first tests whether expert judgments are internally consistent and whether the VLM rubric evaluator agrees with their consensus. The second tests whether DrawAI-Bench tracks the perceived quality and practical usability of artifacts converted to PPTX. The sample spans all four domains, real and AI-generated source images, and the full range of benchmark scores.

\paragraph{Participants and protocol.}
Five experts with at least two years of experience editing vector graphics, presentation slides, diagrams, or scientific figures complete a shared tutorial and a 20-example calibration round that is excluded from analysis. Model, harness, workflow, file provenance, and automatic scores remain hidden, and each expert receives an independently randomized artifact order.

\paragraph{Rubric alignment.}
For each of the 30 rubric-based criteria, all five experts independently assess 40 judgment units, yielding 1,200 units. A unit contains the complete artifact for an artifact-level criterion or an annotated element and its context for an asset-level criterion; Image is excluded because its criteria are rule-based. Experts use the same evidence and question as the VLM evaluator and answer \emph{yes}, \emph{partial}, \emph{no}, or \emph{N/A}, mapped to $1$, $0.5$, $0$, and inapplicability. We form consensus by majority vote, mark a unit N/A when at least three experts do so, and adjudicate only ties after independent ratings are locked. We measure inter-expert consistency with ordinal Krippendorff's $\alpha$ and VLM alignment by exact agreement with the consensus on applicable units.

\paragraph{PPTX usability.}
We convert the same artifacts through one deterministic PPTX pipeline. Each expert opens every PPTX beside its source raster and assigns one score that jointly reflects output quality and usefulness for subsequent editing: 1 denotes an unusable result that must be restarted; 2, major repairs for which restarting is preferable; 3, a usable starting point despite substantial remaining work; 4, a mostly faithful and directly operable result needing limited refinement; and 5, a result ready for reuse with minimal correction. Every artifact receives five independent ratings, for 1,000 ratings in total. We average the ratings per artifact, report the share at or above the usable threshold of 3, and compute Spearman correlations with Fidelity, Editability, and Overall.

\begin{table*}[htbp]
\centering
\caption{\textbf{Human validation of DrawAI-Bench.} Panel (a) reports consistency among the five experts and exact agreement between the VLM evaluator and their consensus. Panel (b) reports practical PPTX usability and its correlation with benchmark scores.}
\label{tab:human-validation-summary}
\begin{minipage}[t]{0.34\textwidth}
\centering
{\RRsans\footnotesize\bfseries (a) Rubric agreement}\par\smallskip
\setlength{\tabcolsep}{4.5pt}
\begin{tabular}{lcc}
\toprule
\rowcolor{DrawAIHeatGray!22}
Taxonomy & Human $\alpha$ & Exact agreement \\
\midrule
Global & \cellcolor{DrawAIHeatBlue!31}0.73 & \cellcolor{DrawAIHeatBlue!34}82.2\% \\
Text & \cellcolor{DrawAIHeatBlue!32}0.75 & \cellcolor{DrawAIHeatBlue!36}87.3\% \\
Formula & \cellcolor{DrawAIHeatBlue!33}0.78 & \cellcolor{DrawAIHeatBlue!38}93.3\% \\
Shape & \cellcolor{DrawAIHeatBlue!33}0.78 & \cellcolor{DrawAIHeatBlue!35}84.2\% \\
Connector & \cellcolor{DrawAIHeatBlue!30}0.69 & \cellcolor{DrawAIHeatBlue!35}83.0\% \\
Table & \cellcolor{DrawAIHeatBlue!29}0.66 & \cellcolor{DrawAIHeatBlue!34}80.1\% \\
\bottomrule

\end{tabular}
\end{minipage}\hfill
\begin{minipage}[t]{0.63\textwidth}
\centering
{\RRsans\footnotesize\bfseries (b) PPTX quality and usability}\par\smallskip
\setlength{\tabcolsep}{2.5pt}
\begin{tabular}{lccccc}
\toprule
\rowcolor{DrawAIHeatGray!22}
Domain & \shortstack{Mean\\score} & \shortstack{Score\\$\geq3$} & \shortstack{Fidelity\\$\rho$} & \shortstack{Editability\\$\rho$} & \shortstack{Overall\\$\rho$} \\
\midrule
Scientific figures & \cellcolor{DrawAIHeatBlue!26}3.26 & \cellcolor{DrawAIHeatBlue!33}78.8\% & \cellcolor{DrawAIHeatBlue!37}0.91 & \cellcolor{DrawAIHeatBlue!38}0.92 & \cellcolor{DrawAIHeatBlue!38}0.94 \\
Presentation slides & \cellcolor{DrawAIHeatBlue!27}3.44 & \cellcolor{DrawAIHeatBlue!35}83.6\% & \cellcolor{DrawAIHeatBlue!38}0.93 & \cellcolor{DrawAIHeatBlue!38}0.91 & \cellcolor{DrawAIHeatBlue!38}0.95 \\
Posters & \cellcolor{DrawAIHeatBlue!23}2.97 & \cellcolor{DrawAIHeatBlue!30}69.6\% & \cellcolor{DrawAIHeatBlue!38}0.94 & \cellcolor{DrawAIHeatBlue!38}0.94 & \cellcolor{DrawAIHeatBlue!38}0.95 \\
Diagrams & \cellcolor{DrawAIHeatBlue!27}3.39 & \cellcolor{DrawAIHeatBlue!36}87.2\% & \cellcolor{DrawAIHeatBlue!36}0.87 & \cellcolor{DrawAIHeatBlue!38}0.92 & \cellcolor{DrawAIHeatBlue!38}0.92 \\
\midrule
\textbf{Overall} & \cellcolor{DrawAIHeatBlue!26}\textbf{3.27} & \cellcolor{DrawAIHeatBlue!34}\textbf{79.8\%} & \cellcolor{DrawAIHeatBlue!37}\textbf{0.88} & \cellcolor{DrawAIHeatBlue!38}\textbf{0.93} & \cellcolor{DrawAIHeatBlue!38}\textbf{0.94} \\
\bottomrule

\end{tabular}
\end{minipage}
\end{table*}

\paragraph{Results.}
Across taxonomy slices, expert agreement ranges from $\alpha=0.66$ to $0.78$, while the VLM exactly matches the five-expert consensus on 80.1--93.3\% of applicable units. Across the 1,000 PPTX ratings, the mean score is 3.27 and 79.8\% are usable. The candidate-level mean human score correlates with Fidelity, Editability, and Overall at $\rho=0.88$, $0.93$, and $0.94$, respectively. Posters are the only domain below the usable mean-score threshold, consistent with their weaker benchmark results. Together, these studies support both the rubric evaluator and the practical relevance of the benchmark's Fidelity--Editability decomposition (Table~\ref{tab:human-validation-summary}).

\begin{figure*}[t!]
\centering
\includegraphics[width=1\textwidth]{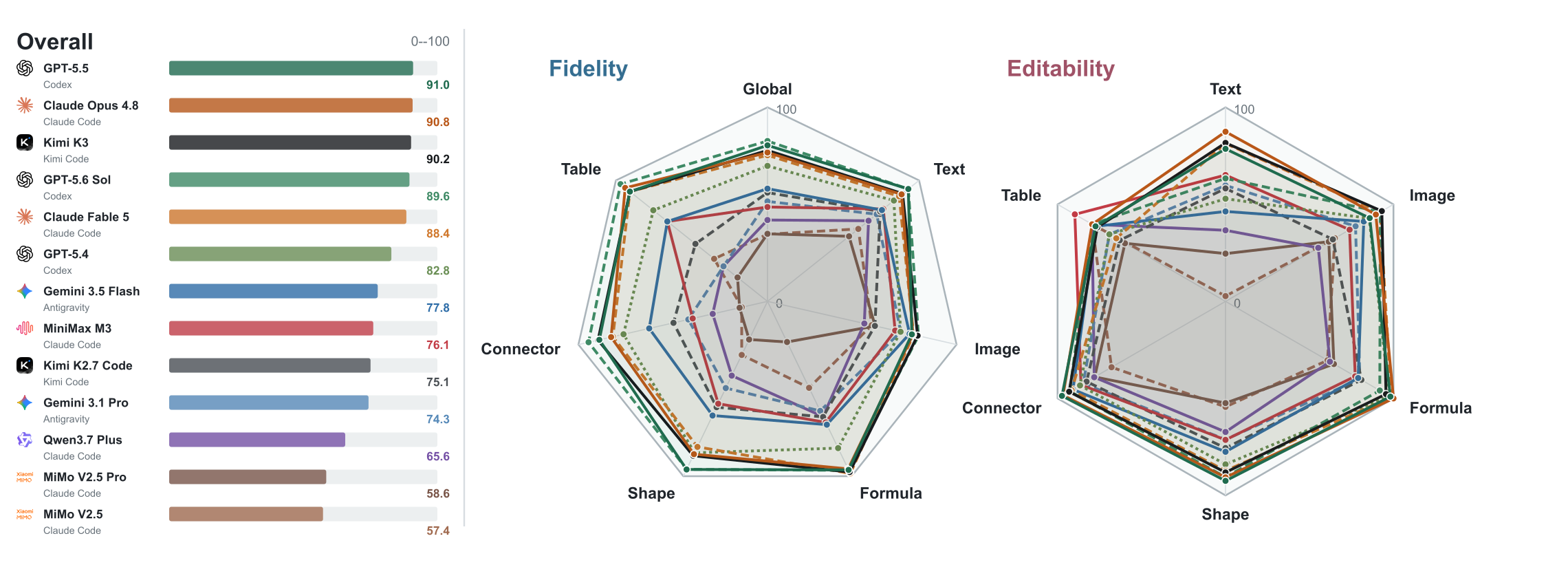}
\caption{\textbf{Model capability profiles under DrawAI-Flow.} The left panel reports Overall scores, while the radar plots separate Fidelity and Editability by asset type for the best-observed model--harness configurations. Chart colors are shared within model families.}
\label{fig:model-results}
\end{figure*}

\subsection{Model Performance}
\label{sec:model-results}
Figure~\ref{fig:model-results} compares each model under its strongest harness. GPT-5.5 leads Overall, GPT-5.6 Sol Fidelity, and Claude Opus 4.8 Editability. This rank reversal shows that visual reproduction and retained authoring structure do not induce a universal model order.

GPT-5.6 Sol makes the distinction concrete, reaching 94.0 Fidelity but 85.1 Editability, and for text 95.7 versus 67.0. It renders text convincingly without reliably recovering reusable text objects, so Fidelity alone overestimates authoring quality. Claude Opus 4.8 reverses this relation and provides the strongest editable text in the leading group. Kimi K3's advantage on editable image objects further shows that similar aggregates can conceal different asset competencies.

The separation also appears across domains. Posters are lowest in both dimensions, indicating broad difficulty with dense mixed compositions. Diagrams rank highest overall, yet Text Editability is only 52.4 while editable images, shapes, and connectors exceed 86.

Tables~\ref{tab:detailed-model-results} and~\ref{tab:domain-results} provide the asset- and domain-level results underlying Figure~\ref{fig:model-results}.

\paragraph{Similar aggregates conceal different compositions.}
The top four configurations span only 1.4 Overall points, yet their Fidelity--Editability balance spans 10.7 points. Across all thirteen models the dimensions are strongly rank-correlated (Spearman's $\rho=0.89$), but ten models score higher on Fidelity than Editability. GPT variants consistently lean toward Fidelity, the two Claude models remain close to parity, and the largest imbalance belongs to \modelmimo{MiMo V2.5}, whose Text Editability falls to 12.5. Editability is therefore the prevailing deficit rather than a symmetric source of variation.

\begin{table*}[htbp]
\centering
\caption{Asset-level DrawAI-Flow results using the strongest observed harness for each model. Darker blue denotes a higher score; gray cells denote an inapplicable asset category.}
\label{tab:detailed-model-results}
\setlength{\tabcolsep}{0.7pt}
\begin{tabular}{ll*{15}{c}}
\toprule
\rowcolor{DrawAIHeatGray!22}
& & \multicolumn{8}{c}{Fidelity} & \multicolumn{7}{c}{Editability} \\
\cmidrule(lr){3-10}\cmidrule(l){11-17}
\rowcolor{DrawAIHeatGray!12}
Model & Harness & Global & Text & Image & Formula & Shape & Connector & Table & Overall & Text & Image & Shape & Connector & Formula & Table & Overall \\
\midrule
\modelopenai{GPT-5.5} & Codex & \cellcolor{DrawAIHeatBlue!34}88.3 & \cellcolor{DrawAIHeatBlue!37}95.7 & \cellcolor{DrawAIHeatBlue!33}85.8 & \cellcolor{DrawAIHeatBlue!37}97.8 & \cellcolor{DrawAIHeatBlue!37}97.7 & \cellcolor{DrawAIHeatBlue!36}93.4 & \cellcolor{DrawAIHeatBlue!36}94.4 & \cellcolor{DrawAIHeatBlue!36}92.8 & \cellcolor{DrawAIHeatBlue!32}80.8 & \cellcolor{DrawAIHeatBlue!34}87.3 & \cellcolor{DrawAIHeatBlue!36}93.2 & \cellcolor{DrawAIHeatBlue!37}97.6 & \cellcolor{DrawAIHeatBlue!37}98.3 & \cellcolor{DrawAIHeatBlue!31}79.6 & \cellcolor{DrawAIHeatBlue!34}89.1 \\
\modelclaude{Claude Opus 4.8} & Claude Code & \cellcolor{DrawAIHeatBlue!33}86.0 & \cellcolor{DrawAIHeatBlue!36}93.1 & \cellcolor{DrawAIHeatBlue!33}86.2 & \cellcolor{DrawAIHeatBlue!37}97.5 & \cellcolor{DrawAIHeatBlue!36}92.4 & \cellcolor{DrawAIHeatBlue!35}89.6 & \cellcolor{DrawAIHeatBlue!37}96.3 & \cellcolor{DrawAIHeatBlue!35}89.9 & \cellcolor{DrawAIHeatBlue!34}88.7 & \cellcolor{DrawAIHeatBlue!35}90.5 & \cellcolor{DrawAIHeatBlue!35}91.6 & \cellcolor{DrawAIHeatBlue!37}97.1 & \cellcolor{DrawAIHeatBlue!38}100.0 & \cellcolor{DrawAIHeatBlue!32}81.5 & \cellcolor{DrawAIHeatBlue!35}91.6 \\
\modelkimi{Kimi K3} & Kimi Code & \cellcolor{DrawAIHeatBlue!34}86.7 & \cellcolor{DrawAIHeatBlue!36}93.4 & \cellcolor{DrawAIHeatBlue!34}87.6 & \cellcolor{DrawAIHeatBlue!38}98.6 & \cellcolor{DrawAIHeatBlue!36}93.0 & \cellcolor{DrawAIHeatBlue!36}93.8 & \cellcolor{DrawAIHeatBlue!36}94.4 & \cellcolor{DrawAIHeatBlue!35}91.1 & \cellcolor{DrawAIHeatBlue!33}83.5 & \cellcolor{DrawAIHeatBlue!36}93.8 & \cellcolor{DrawAIHeatBlue!34}89.3 & \cellcolor{DrawAIHeatBlue!36}93.8 & \cellcolor{DrawAIHeatBlue!37}96.0 & \cellcolor{DrawAIHeatBlue!31}78.7 & \cellcolor{DrawAIHeatBlue!34}89.3 \\
\modelopenai{GPT-5.6 Sol} & Codex & \cellcolor{DrawAIHeatBlue!35}89.6 & \cellcolor{DrawAIHeatBlue!37}95.7 & \cellcolor{DrawAIHeatBlue!34}87.9 & \cellcolor{DrawAIHeatBlue!37}98.0 & \cellcolor{DrawAIHeatBlue!37}97.6 & \cellcolor{DrawAIHeatBlue!37}96.8 & \cellcolor{DrawAIHeatBlue!37}98.2 & \cellcolor{DrawAIHeatBlue!36}94.0 & \cellcolor{DrawAIHeatBlue!27}67.0 & \cellcolor{DrawAIHeatBlue!36}93.0 & \cellcolor{DrawAIHeatBlue!35}91.5 & \cellcolor{DrawAIHeatBlue!36}93.5 & \cellcolor{DrawAIHeatBlue!36}92.7 & \cellcolor{DrawAIHeatBlue!32}81.5 & \cellcolor{DrawAIHeatBlue!33}85.1 \\
\modelclaude{Claude Fable 5} & Claude Code & \cellcolor{DrawAIHeatBlue!33}85.2 & \cellcolor{DrawAIHeatBlue!35}92.1 & \cellcolor{DrawAIHeatBlue!33}86.4 & \cellcolor{DrawAIHeatBlue!38}99.0 & \cellcolor{DrawAIHeatBlue!35}89.9 & \cellcolor{DrawAIHeatBlue!34}89.1 & \cellcolor{DrawAIHeatBlue!36}94.4 & \cellcolor{DrawAIHeatBlue!34}88.8 & \cellcolor{DrawAIHeatBlue!32}83.2 & \cellcolor{DrawAIHeatBlue!36}92.8 & \cellcolor{DrawAIHeatBlue!34}88.6 & \cellcolor{DrawAIHeatBlue!36}92.4 & \cellcolor{DrawAIHeatBlue!38}98.8 & \cellcolor{DrawAIHeatBlue!28}68.5 & \cellcolor{DrawAIHeatBlue!34}88.1 \\
\modelopenai{GPT-5.4} & Codex & \cellcolor{DrawAIHeatBlue!32}81.9 & \cellcolor{DrawAIHeatBlue!35}90.0 & \cellcolor{DrawAIHeatBlue!32}82.2 & \cellcolor{DrawAIHeatBlue!35}90.4 & \cellcolor{DrawAIHeatBlue!35}92.0 & \cellcolor{DrawAIHeatBlue!33}85.7 & \cellcolor{DrawAIHeatBlue!33}85.2 & \cellcolor{DrawAIHeatBlue!34}87.0 & \cellcolor{DrawAIHeatBlue!24}57.6 & \cellcolor{DrawAIHeatBlue!34}87.0 & \cellcolor{DrawAIHeatBlue!33}85.6 & \cellcolor{DrawAIHeatBlue!34}88.0 & \cellcolor{DrawAIHeatBlue!37}97.8 & \cellcolor{DrawAIHeatBlue!29}72.2 & \cellcolor{DrawAIHeatBlue!31}78.7 \\
\modelgemini{Gemini 3.5 Flash} & Antigravity & \cellcolor{DrawAIHeatBlue!30}74.8 & \cellcolor{DrawAIHeatBlue!33}85.3 & \cellcolor{DrawAIHeatBlue!33}85.1 & \cellcolor{DrawAIHeatBlue!32}82.3 & \cellcolor{DrawAIHeatBlue!31}79.2 & \cellcolor{DrawAIHeatBlue!31}77.5 & \cellcolor{DrawAIHeatBlue!31}79.6 & \cellcolor{DrawAIHeatBlue!32}80.5 & \cellcolor{DrawAIHeatBlue!22}51.7 & \cellcolor{DrawAIHeatBlue!33}84.1 & \cellcolor{DrawAIHeatBlue!31}79.8 & \cellcolor{DrawAIHeatBlue!35}91.8 & \cellcolor{DrawAIHeatBlue!32}81.1 & \cellcolor{DrawAIHeatBlue!31}79.6 & \cellcolor{DrawAIHeatBlue!30}75.0 \\
\modelminimax{MiniMax M3} & Claude Code & \cellcolor{DrawAIHeatBlue!28}69.2 & \cellcolor{DrawAIHeatBlue!33}85.8 & \cellcolor{DrawAIHeatBlue!32}80.5 & \cellcolor{DrawAIHeatBlue!32}81.5 & \cellcolor{DrawAIHeatBlue!30}75.1 & \cellcolor{DrawAIHeatBlue!26}63.7 & \cellcolor{DrawAIHeatBlue!31}79.6 & \cellcolor{DrawAIHeatBlue!30}75.7 & \cellcolor{DrawAIHeatBlue!28}68.6 & \cellcolor{DrawAIHeatBlue!30}76.5 & \cellcolor{DrawAIHeatBlue!29}74.2 & \cellcolor{DrawAIHeatBlue!34}87.5 & \cellcolor{DrawAIHeatBlue!31}79.5 & \cellcolor{DrawAIHeatBlue!35}90.7 & \cellcolor{DrawAIHeatBlue!30}76.5 \\
\modelkimi{Kimi K2.7 Code} & Kimi Code & \cellcolor{DrawAIHeatBlue!29}73.6 & \cellcolor{DrawAIHeatBlue!33}84.6 & \cellcolor{DrawAIHeatBlue!29}74.1 & \cellcolor{DrawAIHeatBlue!31}79.7 & \cellcolor{DrawAIHeatBlue!30}76.3 & \cellcolor{DrawAIHeatBlue!28}69.8 & \cellcolor{DrawAIHeatBlue!28}68.5 & \cellcolor{DrawAIHeatBlue!30}76.6 & \cellcolor{DrawAIHeatBlue!26}62.4 & \cellcolor{DrawAIHeatBlue!27}67.5 & \cellcolor{DrawAIHeatBlue!31}78.0 & \cellcolor{DrawAIHeatBlue!33}84.5 & \cellcolor{DrawAIHeatBlue!32}82.8 & \cellcolor{DrawAIHeatBlue!27}66.7 & \cellcolor{DrawAIHeatBlue!29}73.6 \\
\modelgemini{Gemini 3.1 Pro} & Antigravity & \cellcolor{DrawAIHeatBlue!28}70.9 & \cellcolor{DrawAIHeatBlue!32}83.2 & \cellcolor{DrawAIHeatBlue!32}82.3 & \cellcolor{DrawAIHeatBlue!31}77.6 & \cellcolor{DrawAIHeatBlue!28}69.8 & \cellcolor{DrawAIHeatBlue!26}65.1 & \cellcolor{DrawAIHeatBlue!24}57.4 & \cellcolor{DrawAIHeatBlue!29}74.1 & \cellcolor{DrawAIHeatBlue!26}63.8 & \cellcolor{DrawAIHeatBlue!31}79.7 & \cellcolor{DrawAIHeatBlue!29}74.1 & \cellcolor{DrawAIHeatBlue!33}86.2 & \cellcolor{DrawAIHeatBlue!32}82.0 & \cellcolor{DrawAIHeatBlue!29}72.2 & \cellcolor{DrawAIHeatBlue!30}74.6 \\
\modelqwen{Qwen3.7 Plus} & Claude Code & \cellcolor{DrawAIHeatBlue!27}65.2 & \cellcolor{DrawAIHeatBlue!31}80.0 & \cellcolor{DrawAIHeatBlue!28}70.7 & \cellcolor{DrawAIHeatBlue!31}79.5 & \cellcolor{DrawAIHeatBlue!27}65.5 & \cellcolor{DrawAIHeatBlue!24}57.4 & \cellcolor{DrawAIHeatBlue!24}57.4 & \cellcolor{DrawAIHeatBlue!28}68.4 & \cellcolor{DrawAIHeatBlue!19}43.0 & \cellcolor{DrawAIHeatBlue!25}59.7 & \cellcolor{DrawAIHeatBlue!28}70.6 & \cellcolor{DrawAIHeatBlue!32}80.4 & \cellcolor{DrawAIHeatBlue!27}65.9 & \cellcolor{DrawAIHeatBlue!32}81.5 & \cellcolor{DrawAIHeatBlue!26}62.9 \\
\modelmimo{MiMo V2.5 Pro} & Claude Code & \cellcolor{DrawAIHeatBlue!25}60.9 & \cellcolor{DrawAIHeatBlue!29}72.3 & \cellcolor{DrawAIHeatBlue!29}74.2 & \cellcolor{DrawAIHeatBlue!23}53.9 & \cellcolor{DrawAIHeatBlue!23}53.1 & \cellcolor{DrawAIHeatBlue!21}48.9 & \cellcolor{DrawAIHeatBlue!22}51.9 & \cellcolor{DrawAIHeatBlue!25}61.0 & \cellcolor{DrawAIHeatBlue!16}32.2 & \cellcolor{DrawAIHeatBlue!27}65.3 & \cellcolor{DrawAIHeatBlue!24}57.2 & \cellcolor{DrawAIHeatBlue!31}80.1 & \cellcolor{DrawAIHeatBlue!27}68.1 & \cellcolor{DrawAIHeatBlue!26}63.9 & \cellcolor{DrawAIHeatBlue!24}56.1 \\
\modelmimo{MiMo V2.5} & Claude Code & \cellcolor{DrawAIHeatBlue!25}60.8 & \cellcolor{DrawAIHeatBlue!30}75.9 & \cellcolor{DrawAIHeatBlue!29}73.1 & \cellcolor{DrawAIHeatBlue!28}69.7 & \cellcolor{DrawAIHeatBlue!24}58.4 & \cellcolor{DrawAIHeatBlue!21}48.2 & \cellcolor{DrawAIHeatBlue!25}61.1 & \cellcolor{DrawAIHeatBlue!26}63.8 & \cellcolor{DrawAIHeatBlue!9}12.5 & \cellcolor{DrawAIHeatBlue!28}68.6 & \cellcolor{DrawAIHeatBlue!24}58.9 & \cellcolor{DrawAIHeatBlue!28}71.1 & \cellcolor{DrawAIHeatBlue!26}64.7 & \cellcolor{DrawAIHeatBlue!32}81.5 & \cellcolor{DrawAIHeatBlue!22}50.9 \\
\bottomrule
\end{tabular}
\end{table*}

\paragraph{No model dominates at the asset level.}
The seven Fidelity categories are led by three models and the six Editability categories by four. Text is the weakest Editability category for nine of thirteen models: its panel mean is 61.2, compared with 86.7 for Text Fidelity. Once editable text is strong, native table grids become the next bottleneck. Sibling variants can also exchange strengths: \modelopenai{GPT-5.6 Sol} exceeds \modelopenai{GPT-5.5} on editable image objects but trails it by 13.8 points on editable text.

\begin{table*}[htbp]
\centering
\caption{Domain-level results pooled across scored DrawAI-Flow configurations. Darker blue denotes a higher score; gray cells denote asset categories not represented in that domain.}
\label{tab:domain-results}
\setlength{\tabcolsep}{2.0pt}
\begin{tabular}{l*{15}{c}}
\toprule
\rowcolor{DrawAIHeatGray!22}
& \multicolumn{8}{c}{Fidelity} & \multicolumn{7}{c}{Editability} \\
\cmidrule(lr){2-9}\cmidrule(l){10-16}
\rowcolor{DrawAIHeatGray!12}
Domain & Global & Text & Image & Formula & Shape & Connector & Table & Overall & Text & Image & Shape & Connector & Formula & Table & Overall \\
\midrule
Diagrams & \cellcolor{DrawAIHeatBlue!30}75.7 & \cellcolor{DrawAIHeatBlue!33}83.7 & \cellcolor{DrawAIHeatBlue!36}92.8 & \cellcolor{DrawAIHeatGray!28}\textit{N/A} & \cellcolor{DrawAIHeatBlue!32}80.5 & \cellcolor{DrawAIHeatBlue!29}72.4 & \cellcolor{DrawAIHeatGray!28}\textit{N/A} & \cellcolor{DrawAIHeatBlue!31}78.8 & \cellcolor{DrawAIHeatBlue!22}52.4 & \cellcolor{DrawAIHeatBlue!35}90.9 & \cellcolor{DrawAIHeatBlue!34}86.4 & \cellcolor{DrawAIHeatBlue!35}90.2 & \cellcolor{DrawAIHeatGray!28}\textit{N/A} & \cellcolor{DrawAIHeatGray!28}\textit{N/A} & \cellcolor{DrawAIHeatBlue!30}77.1 \\
Posters & \cellcolor{DrawAIHeatBlue!27}65.9 & \cellcolor{DrawAIHeatBlue!32}80.9 & \cellcolor{DrawAIHeatBlue!29}72.7 & \cellcolor{DrawAIHeatGray!28}\textit{N/A} & \cellcolor{DrawAIHeatBlue!26}65.1 & \cellcolor{DrawAIHeatBlue!26}62.8 & \cellcolor{DrawAIHeatBlue!29}71.7 & \cellcolor{DrawAIHeatBlue!28}70.6 & \cellcolor{DrawAIHeatBlue!26}62.6 & \cellcolor{DrawAIHeatBlue!28}71.1 & \cellcolor{DrawAIHeatBlue!24}58.7 & \cellcolor{DrawAIHeatBlue!28}69.9 & \cellcolor{DrawAIHeatGray!28}\textit{N/A} & \cellcolor{DrawAIHeatBlue!28}69.6 & \cellcolor{DrawAIHeatBlue!27}65.6 \\
Presentation slides & \cellcolor{DrawAIHeatBlue!30}76.9 & \cellcolor{DrawAIHeatBlue!33}85.0 & \cellcolor{DrawAIHeatBlue!31}79.5 & \cellcolor{DrawAIHeatGray!28}\textit{N/A} & \cellcolor{DrawAIHeatBlue!30}75.3 & \cellcolor{DrawAIHeatBlue!29}72.3 & \cellcolor{DrawAIHeatBlue!33}84.9 & \cellcolor{DrawAIHeatBlue!31}78.1 & \cellcolor{DrawAIHeatBlue!26}63.7 & \cellcolor{DrawAIHeatBlue!30}75.5 & \cellcolor{DrawAIHeatBlue!30}75.5 & \cellcolor{DrawAIHeatBlue!33}85.1 & \cellcolor{DrawAIHeatGray!28}\textit{N/A} & \cellcolor{DrawAIHeatBlue!38}100.0 & \cellcolor{DrawAIHeatBlue!29}72.8 \\
Scientific figures & \cellcolor{DrawAIHeatBlue!29}74.0 & \cellcolor{DrawAIHeatBlue!33}84.1 & \cellcolor{DrawAIHeatBlue!32}82.9 & \cellcolor{DrawAIHeatBlue!31}79.9 & \cellcolor{DrawAIHeatBlue!30}74.5 & \cellcolor{DrawAIHeatBlue!29}72.1 & \cellcolor{DrawAIHeatBlue!29}73.5 & \cellcolor{DrawAIHeatBlue!31}77.5 & \cellcolor{DrawAIHeatBlue!24}57.1 & \cellcolor{DrawAIHeatBlue!31}79.6 & \cellcolor{DrawAIHeatBlue!31}77.6 & \cellcolor{DrawAIHeatBlue!34}88.0 & \cellcolor{DrawAIHeatBlue!32}82.1 & \cellcolor{DrawAIHeatBlue!29}72.1 & \cellcolor{DrawAIHeatBlue!30}76.1 \\
\bottomrule
\end{tabular}
\end{table*}

\paragraph{Domain difficulty is chiefly an Editability phenomenon.}
The domain-level Overall scores order Diagrams (78.0), Scientific figures (76.8), Presentation slides (75.5), and Posters (68.1). Text Fidelity varies by only 4.1 points across domains, whereas Editability categories vary by as much as 27.7 points. Diagrams obtain the strongest Overall score despite the weakest Text Editability (52.4), while Posters retain more editable text but degrade most on Global Fidelity, Shape Fidelity, and Shape Editability. The human study in Section~\ref{sec:human-validation} mirrors this ordering: Posters are the only domain below the usable mean-score threshold.

\subsection{Harness Effects}
\label{sec:harness-results}

\begin{figure*}[tbp]
\centering
\includegraphics[width=0.5\textwidth]{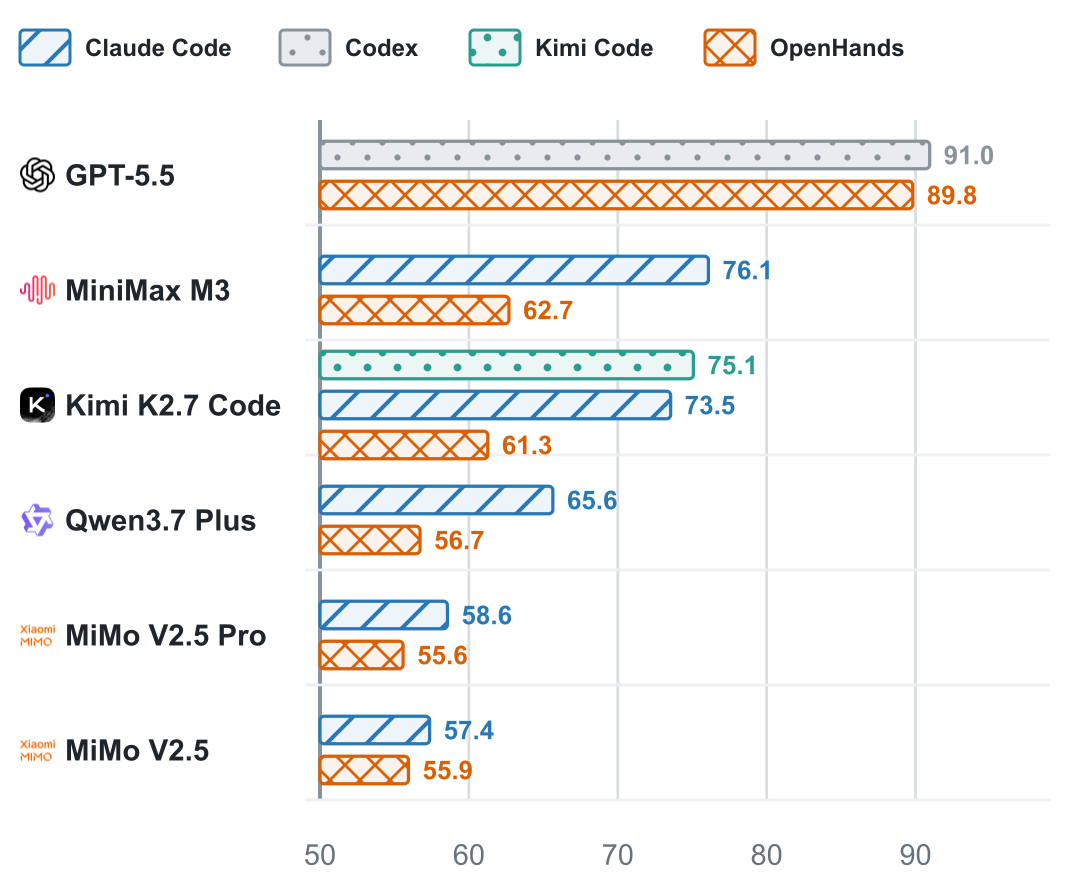}
\caption{\textbf{Effect of the execution harness.} Overall scores for models evaluated with multiple harnesses.}
\label{fig:harness-impact}
\end{figure*}

Figure~\ref{fig:harness-impact} compares models evaluated with at least two harnesses and keeps every comparison within model and source image. Claude Code outperforms OpenHands for every model evaluated with both harnesses. The Overall gains are substantial for Kimi K2.7 Code, MiniMax M3, and Qwen3.7 Plus, but smaller for the two MiMo variants. Provider-aligned pairings also attain the best-observed result where a direct comparison is available, where GPT-5.5 performs best with Codex, and Kimi K2.7 Code with Kimi Code. These results validate that matching the model to an effective, often provider-aligned harness is more important than selecting a universally best harness.

Figure~\ref{fig:harness-impact} reports the absolute Overall results, while Table~\ref{tab:paired-harness-effects} reports matched-image effects relative to OpenHands.

\paragraph{Sensitivity peaks in the middle of the capability range.}
The absolute and paired views agree to within 1.7 Overall points. Harness-induced ranges reach 13.4 points for \modelminimax{MiniMax M3} and 13.8 for \modelkimi{Kimi K2.7 Code}, but only 1.2 for \modelopenai{GPT-5.5}. Relative to OpenHands, gains are largest for the mid-capability models and small at both ends: the strongest model can compensate for weaker scaffolding, while the weakest models cannot fully exploit a stronger harness.

\begin{table*}[htbp]
\centering
\caption{Within-image harness effects relative to OpenHands, reported as mean $\pm$ sample standard deviation across paired images. Blue cells favor the alternative harness and red cells favor OpenHands.}
\label{tab:paired-harness-effects}
\setlength{\tabcolsep}{5.0pt}
\begin{tabular}{lllrrr}
\toprule
\rowcolor{DrawAIHeatGray!22}
Model & Alternative & Reference & $\Delta$ Fidelity & $\Delta$ Editability & $\Delta$ Overall \\
\midrule
\modelkimi{Kimi K2.7 Code} & Kimi Code & OpenHands & \cellcolor{DrawAIHeatBlue!27}+17.8 $\pm$ 15.1 & \cellcolor{DrawAIHeatBlue!20}+11.9 $\pm$ 20.7 & \cellcolor{DrawAIHeatBlue!24}+14.9 $\pm$ 15.8 \\
\modelkimi{Kimi K2.7 Code} & Claude Code & OpenHands & \cellcolor{DrawAIHeatBlue!25}+16.3 $\pm$ 11.6 & \cellcolor{DrawAIHeatBlue!19}+11.5 $\pm$ 19.5 & \cellcolor{DrawAIHeatBlue!22}+13.9 $\pm$ 12.8 \\
\modelminimax{MiniMax M3} & Claude Code & OpenHands & \cellcolor{DrawAIHeatBlue!22}+13.8 $\pm$ 13.6 & \cellcolor{DrawAIHeatBlue!21}+13.1 $\pm$ 17.2 & \cellcolor{DrawAIHeatBlue!22}+13.5 $\pm$ 13.1 \\
\modelqwen{Qwen3.7 Plus} & Claude Code & OpenHands & \cellcolor{DrawAIHeatBlue!16}+8.6 $\pm$ 10.9 & \cellcolor{DrawAIHeatBlue!17}+9.2 $\pm$ 18.3 & \cellcolor{DrawAIHeatBlue!16}+8.9 $\pm$ 12.8 \\
\modelmimo{MiMo V2.5 Pro} & Claude Code & OpenHands & \cellcolor{DrawAIHeatBlue!11}+4.6 $\pm$ 14.5 & \cellcolor{DrawAIHeatBlue!7}+1.4 $\pm$ 19.4 & \cellcolor{DrawAIHeatBlue!9}+3.0 $\pm$ 15.9 \\
\modelmimo{MiMo V2.5} & Claude Code & OpenHands & \cellcolor{DrawAIHeatBlue!15}+8.2 $\pm$ 15.4 & \cellcolor{DrawAIHeatRed!12}-5.3 $\pm$ 22.0 & \cellcolor{DrawAIHeatBlue!7}+1.4 $\pm$ 17.1 \\
\modelopenai{GPT-5.5} & Codex & OpenHands & \cellcolor{DrawAIHeatBlue!6}+0.9 $\pm$ 4.3 & \cellcolor{DrawAIHeatBlue!7}+1.3 $\pm$ 12.5 & \cellcolor{DrawAIHeatBlue!6}+1.1 $\pm$ 6.8 \\
\bottomrule
\end{tabular}
\end{table*}

\paragraph{Harnesses change the error profile as well as the score.}
Paired standard deviations remain large, so a favorable harness still loses on some images. For \modelmimo{MiMo V2.5}, Claude Code raises Fidelity by 8.2 points relative to OpenHands while lowering Editability by 5.3, turning a balanced profile into a Fidelity-heavy one. Harness choice therefore participates in the Fidelity--Editability tension rather than merely scaling quality.

\subsection{Workflow Effects}
\label{sec:method-results}

\begin{figure*}[tbp]
\centering
\includegraphics[width=0.5\textwidth]{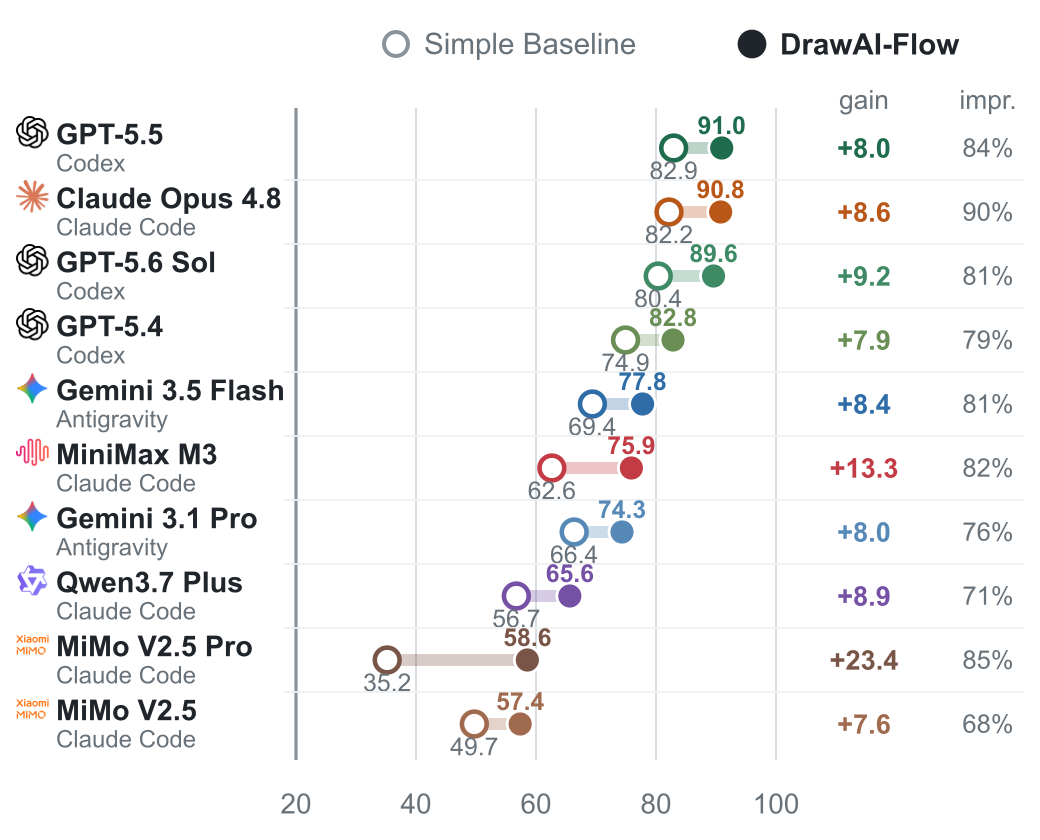}
\caption{\textbf{Effect of DrawAI-Flow.} Paired Overall scores compare direct reconstruction with DrawAI-Flow under the same model, harness, and image; the right columns report mean gain and the percentage of images improved.}
\label{fig:method-interaction}
\end{figure*}

Figure~\ref{fig:method-interaction} separates direct capability from workflow gain. DrawAI-Flow improves mean Overall for every displayed configuration, including strong direct agents, but gain size is not monotonic in the final score. Across all paired outputs, Fidelity rises from 75.7 to 78.7, whereas Editability rises from 56.4 to 74.0, showing that the workflow primarily closes the structural gap.

The asset breakdown explains why Editability moves more. DrawAI-Flow produces its largest gains on editable text and image objects, followed by formulas and connectors. Among Fidelity categories, only images improve comparably; formula, shape, and connector appearance regress slightly. This pattern indicates that explicit planning and materialization turn recognized content into separable authoring objects, but do not uniformly solve geometric appearance. The intervention also remains instance-dependent: only 68--90\% of paired images improve across settings, and most retain a negative lower tail.

Tables~\ref{tab:pooled-workflow-asset-effects}--\ref{tab:workflow-asset-gains} expand the aggregate comparison to pooled assets, individual model--harness settings, per-image gain distributions, and model-specific asset effects.
\begin{table*}[t]
\centering
\caption{Pooled asset-level effect of DrawAI-Flow relative to the Simple Baseline, reported as mean $\pm$ sample standard deviation across paired images. Blue cells indicate improvement and red cells regression.}
\label{tab:pooled-workflow-asset-effects}
\setlength{\tabcolsep}{10pt}
\begin{tabular}{llr}
\toprule
\rowcolor{DrawAIHeatGray!22}
Dimension & Asset & DrawAI-Flow gain \\
\midrule
Fidelity & Global & \cellcolor{DrawAIHeatBlue!6}+1.6 $\pm$ 14.6 \\
Fidelity & Text & \cellcolor{DrawAIHeatBlue!8}+4.2 $\pm$ 18.0 \\
Fidelity & Image & \cellcolor{DrawAIHeatBlue!20}+21.4 $\pm$ 24.5 \\
Fidelity & Formula & \cellcolor{DrawAIHeatRed!9}-6.0 $\pm$ 22.9 \\
Fidelity & Shape & \cellcolor{DrawAIHeatRed!7}-2.3 $\pm$ 20.2 \\
Fidelity & Connector & \cellcolor{DrawAIHeatRed!6}-2.1 $\pm$ 25.0 \\
Fidelity & Table & \cellcolor{DrawAIHeatBlue!6}+0.7 $\pm$ 31.3 \\
Editability & Text & \cellcolor{DrawAIHeatBlue!27}+30.4 $\pm$ 34.3 \\
Editability & Image & \cellcolor{DrawAIHeatBlue!24}+26.9 $\pm$ 49.4 \\
Editability & Shape & \cellcolor{DrawAIHeatBlue!9}+5.9 $\pm$ 22.6 \\
Editability & Connector & \cellcolor{DrawAIHeatBlue!14}+12.6 $\pm$ 34.2 \\
Editability & Formula & \cellcolor{DrawAIHeatBlue!20}+20.8 $\pm$ 36.8 \\
Editability & Table & \cellcolor{DrawAIHeatBlue!9}+6.2 $\pm$ 35.3 \\
\bottomrule
\end{tabular}
\end{table*}

\paragraph{The workflow converts recognized content into structure.}
Across paired outputs, DrawAI-Flow raises Editability by 17.6 points but Fidelity by 3.0. Text Editability makes the largest single movement, from 26.1 to 56.5. Image is the only category that improves substantially on both dimensions, reflecting the value of planned localized assets. Formula, shape, and connector Fidelity regress slightly while their Editability improves, exposing a deliberate shift toward native, manipulable primitives.

\begin{table*}[t]
\centering
\caption{Complete paired workflow comparison over unique model--harness settings. Duplicate execution routes within a harness are collapsed to the higher-scoring DrawAI-Flow result. Darker blue denotes higher absolute quality; blue and red gain cells denote improvements and regressions, respectively.}
\label{tab:paired-workflow-ablation}
\setlength{\tabcolsep}{5pt}
\begin{tabular}{llrrrrr}
\toprule
\rowcolor{DrawAIHeatGray!22}
Model & Harness & Simple Baseline & DrawAI-Flow & $\Delta$ Fidelity & $\Delta$ Editability & $\Delta$ Overall \\
\midrule
\modelopenai{GPT-5.5} & Codex & \cellcolor{DrawAIHeatBlue!32}82.9 & \cellcolor{DrawAIHeatBlue!35}\textbf{91.0} & \cellcolor{DrawAIHeatBlue!9}+3.3 & \cellcolor{DrawAIHeatBlue!16}+12.7 & \cellcolor{DrawAIHeatBlue!12}\textbf{+8.0} \\
\modelclaude{Claude Opus 4.8} & Claude Code & \cellcolor{DrawAIHeatBlue!32}82.2 & \cellcolor{DrawAIHeatBlue!35}\textbf{90.8} & \cellcolor{DrawAIHeatBlue!9}+3.2 & \cellcolor{DrawAIHeatBlue!17}+14.0 & \cellcolor{DrawAIHeatBlue!13}\textbf{+8.6} \\
\modelopenai{GPT-5.5} & OpenHands & \cellcolor{DrawAIHeatBlue!31}77.7 & \cellcolor{DrawAIHeatBlue!35}\textbf{89.8} & \cellcolor{DrawAIHeatBlue!11}+5.9 & \cellcolor{DrawAIHeatBlue!20}+18.4 & \cellcolor{DrawAIHeatBlue!15}\textbf{+12.1} \\
\modelopenai{GPT-5.6 Sol} & Codex & \cellcolor{DrawAIHeatBlue!32}80.4 & \cellcolor{DrawAIHeatBlue!35}\textbf{89.6} & \cellcolor{DrawAIHeatBlue!10}+3.7 & \cellcolor{DrawAIHeatBlue!17}+14.7 & \cellcolor{DrawAIHeatBlue!13}\textbf{+9.2} \\
\modelopenai{GPT-5.4} & Codex & \cellcolor{DrawAIHeatBlue!30}74.9 & \cellcolor{DrawAIHeatBlue!32}\textbf{82.8} & \cellcolor{DrawAIHeatBlue!9}+3.3 & \cellcolor{DrawAIHeatBlue!16}+12.5 & \cellcolor{DrawAIHeatBlue!12}\textbf{+7.9} \\
\modelgemini{Gemini 3.5 Flash} & Antigravity & \cellcolor{DrawAIHeatBlue!28}69.4 & \cellcolor{DrawAIHeatBlue!31}\textbf{77.8} & \cellcolor{DrawAIHeatBlue!9}+3.4 & \cellcolor{DrawAIHeatBlue!16}+13.3 & \cellcolor{DrawAIHeatBlue!13}\textbf{+8.4} \\
\modelminimax{MiniMax M3} & Claude Code & \cellcolor{DrawAIHeatBlue!26}62.6 & \cellcolor{DrawAIHeatBlue!30}\textbf{75.9} & \cellcolor{DrawAIHeatBlue!9}+2.4 & \cellcolor{DrawAIHeatBlue!23}+24.1 & \cellcolor{DrawAIHeatBlue!16}\textbf{+13.3} \\
\modelgemini{Gemini 3.1 Pro} & Antigravity & \cellcolor{DrawAIHeatBlue!27}66.4 & \cellcolor{DrawAIHeatBlue!30}\textbf{74.3} & \cellcolor{DrawAIHeatRed!7}-0.5 & \cellcolor{DrawAIHeatBlue!18}+16.4 & \cellcolor{DrawAIHeatBlue!12}\textbf{+8.0} \\
\modelqwen{Qwen3.7 Plus} & Claude Code & \cellcolor{DrawAIHeatBlue!24}56.7 & \cellcolor{DrawAIHeatBlue!27}\textbf{65.6} & \cellcolor{DrawAIHeatBlue!8}+0.9 & \cellcolor{DrawAIHeatBlue!19}+16.9 & \cellcolor{DrawAIHeatBlue!13}\textbf{+8.9} \\
\modelminimax{MiniMax M3} & OpenHands & \cellcolor{DrawAIHeatBlue!16}33.3 & \cellcolor{DrawAIHeatBlue!26}\textbf{62.7} & \cellcolor{DrawAIHeatBlue!15}+12.3 & \cellcolor{DrawAIHeatBlue!39}+46.6 & \cellcolor{DrawAIHeatBlue!27}\textbf{+29.4} \\
\modelmimo{MiMo V2.5 Pro} & Claude Code & \cellcolor{DrawAIHeatBlue!17}35.2 & \cellcolor{DrawAIHeatBlue!24}\textbf{58.6} & \cellcolor{DrawAIHeatBlue!13}+8.4 & \cellcolor{DrawAIHeatBlue!33}+38.3 & \cellcolor{DrawAIHeatBlue!23}\textbf{+23.4} \\
\modelmimo{MiMo V2.5} & Claude Code & \cellcolor{DrawAIHeatBlue!21}49.7 & \cellcolor{DrawAIHeatBlue!24}\textbf{57.4} & \cellcolor{DrawAIHeatBlue!8}+1.4 & \cellcolor{DrawAIHeatBlue!16}+13.8 & \cellcolor{DrawAIHeatBlue!12}\textbf{+7.6} \\
\modelmimo{MiMo V2.5} & OpenHands & \cellcolor{DrawAIHeatBlue!13}23.1 & \cellcolor{DrawAIHeatBlue!23}\textbf{55.9} & \cellcolor{DrawAIHeatBlue!23}+23.4 & \cellcolor{DrawAIHeatBlue!36}+42.3 & \cellcolor{DrawAIHeatBlue!29}\textbf{+32.9} \\
\modelmimo{MiMo V2.5 Pro} & OpenHands & \cellcolor{DrawAIHeatBlue!12}22.4 & \cellcolor{DrawAIHeatBlue!23}\textbf{55.6} & \cellcolor{DrawAIHeatBlue!22}+22.6 & \cellcolor{DrawAIHeatBlue!37}+43.7 & \cellcolor{DrawAIHeatBlue!30}\textbf{+33.2} \\
\bottomrule
\end{tabular}
\end{table*}

\paragraph{The workflow lifts the floor and partially substitutes for the harness.}
All fourteen model--harness settings improve, and Editability gain exceeds Fidelity gain in every setting. Gain is strongly anticorrelated with baseline strength (Pearson $r=-0.89$): the workflow raises the panel minimum from 22.4 to 55.6 and reduces the cross-setting standard deviation from 22.0 to 14.0. For models tested under two harnesses, it shrinks the harness-induced Overall gap by 55--94\%, although the best results still require a strong model--harness pairing.

\begin{table*}[t]
\centering
\caption{Distribution of per-image Overall gains from DrawAI-Flow. Blue and red cells denote positive and negative gains; the final column reports the share of paired images improved by the workflow.}
\label{tab:workflow-gain-distributions}
\setlength{\tabcolsep}{5pt}
\begin{tabular}{llrrrrr}
\toprule
\rowcolor{DrawAIHeatGray!22}
Model & Harness & Mean $\Delta$ Overall & 10th percentile & Median & 90th percentile & Improved images (\%) \\
\midrule
\modelopenai{GPT-5.5} & Codex & \cellcolor{DrawAIHeatBlue!12}\textbf{+8.0} & \cellcolor{DrawAIHeatRed!9}-3.6 & \cellcolor{DrawAIHeatBlue!12}+7.5 & \cellcolor{DrawAIHeatBlue!21}+21.1 & \cellcolor{DrawAIHeatBlue!33}\textbf{83.8} \\
\modelclaude{Claude Opus 4.8} & Claude Code & \cellcolor{DrawAIHeatBlue!13}\textbf{+8.6} & \cellcolor{DrawAIHeatGray!24}0.0 & \cellcolor{DrawAIHeatBlue!12}+6.7 & \cellcolor{DrawAIHeatBlue!21}+20.4 & \cellcolor{DrawAIHeatBlue!35}\textbf{90.0} \\
\modelopenai{GPT-5.5} & OpenHands & \cellcolor{DrawAIHeatBlue!15}\textbf{+12.1} & \cellcolor{DrawAIHeatRed!8}-1.0 & \cellcolor{DrawAIHeatBlue!16}+13.6 & \cellcolor{DrawAIHeatBlue!25}+26.1 & \cellcolor{DrawAIHeatBlue!34}\textbf{87.5} \\
\modelopenai{GPT-5.6 Sol} & Codex & \cellcolor{DrawAIHeatBlue!13}\textbf{+9.2} & \cellcolor{DrawAIHeatRed!9}-3.3 & \cellcolor{DrawAIHeatBlue!13}+8.8 & \cellcolor{DrawAIHeatBlue!22}+22.2 & \cellcolor{DrawAIHeatBlue!32}\textbf{81.3} \\
\modelopenai{GPT-5.4} & Codex & \cellcolor{DrawAIHeatBlue!12}\textbf{+7.9} & \cellcolor{DrawAIHeatRed!10}-3.9 & \cellcolor{DrawAIHeatBlue!12}+7.9 & \cellcolor{DrawAIHeatBlue!21}+20.7 & \cellcolor{DrawAIHeatBlue!31}\textbf{78.8} \\
\modelgemini{Gemini 3.5 Flash} & Antigravity & \cellcolor{DrawAIHeatBlue!13}\textbf{+8.4} & \cellcolor{DrawAIHeatRed!10}-5.1 & \cellcolor{DrawAIHeatBlue!11}+6.6 & \cellcolor{DrawAIHeatBlue!22}+22.6 & \cellcolor{DrawAIHeatBlue!32}\textbf{81.3} \\
\modelminimax{MiniMax M3} & Claude Code & \cellcolor{DrawAIHeatBlue!16}\textbf{+13.3} & \cellcolor{DrawAIHeatRed!12}-7.4 & \cellcolor{DrawAIHeatBlue!18}+15.5 & \cellcolor{DrawAIHeatBlue!27}+29.4 & \cellcolor{DrawAIHeatBlue!32}\textbf{82.0} \\
\modelgemini{Gemini 3.1 Pro} & Antigravity & \cellcolor{DrawAIHeatBlue!12}\textbf{+8.0} & \cellcolor{DrawAIHeatRed!13}-9.1 & \cellcolor{DrawAIHeatBlue!13}+8.4 & \cellcolor{DrawAIHeatBlue!22}+22.3 & \cellcolor{DrawAIHeatBlue!30}\textbf{76.3} \\
\modelqwen{Qwen3.7 Plus} & Claude Code & \cellcolor{DrawAIHeatBlue!13}\textbf{+8.9} & \cellcolor{DrawAIHeatRed!10}-4.8 & \cellcolor{DrawAIHeatBlue!11}+6.4 & \cellcolor{DrawAIHeatBlue!25}+26.6 & \cellcolor{DrawAIHeatBlue!29}\textbf{71.3} \\
\modelminimax{MiniMax M3} & OpenHands & \cellcolor{DrawAIHeatBlue!27}\textbf{+29.4} & \cellcolor{DrawAIHeatRed!10}-4.9 & \cellcolor{DrawAIHeatBlue!28}+30.4 & \cellcolor{DrawAIHeatBlue!41}+65.5 & \cellcolor{DrawAIHeatBlue!33}\textbf{85.0} \\
\modelmimo{MiMo V2.5 Pro} & Claude Code & \cellcolor{DrawAIHeatBlue!23}\textbf{+23.4} & \cellcolor{DrawAIHeatRed!13}-8.6 & \cellcolor{DrawAIHeatBlue!25}+26.7 & \cellcolor{DrawAIHeatBlue!41}+51.0 & \cellcolor{DrawAIHeatBlue!33}\textbf{85.0} \\
\modelmimo{MiMo V2.5} & Claude Code & \cellcolor{DrawAIHeatBlue!12}\textbf{+7.6} & \cellcolor{DrawAIHeatRed!13}-8.6 & \cellcolor{DrawAIHeatBlue!14}+9.9 & \cellcolor{DrawAIHeatBlue!23}+23.6 & \cellcolor{DrawAIHeatBlue!27}\textbf{67.5} \\
\modelmimo{MiMo V2.5} & OpenHands & \cellcolor{DrawAIHeatBlue!29}\textbf{+32.9} & \cellcolor{DrawAIHeatBlue!8}+1.0 & \cellcolor{DrawAIHeatBlue!30}+33.9 & \cellcolor{DrawAIHeatBlue!41}+61.5 & \cellcolor{DrawAIHeatBlue!35}\textbf{90.0} \\
\modelmimo{MiMo V2.5 Pro} & OpenHands & \cellcolor{DrawAIHeatBlue!30}\textbf{+33.2} & \cellcolor{DrawAIHeatRed!9}-3.0 & \cellcolor{DrawAIHeatBlue!30}+34.0 & \cellcolor{DrawAIHeatBlue!41}+62.1 & \cellcolor{DrawAIHeatBlue!34}\textbf{88.8} \\
\bottomrule
\end{tabular}\end{table*}

\begin{table*}[t]
\centering
\caption{Mean per-model asset-level effect of DrawAI-Flow relative to the Simple Baseline. Blue cells indicate improvement and red cells regression.}
\label{tab:workflow-asset-gains}
\setlength{\tabcolsep}{2.0pt}
\begin{tabular}{ll*{13}{r}}
\toprule
\rowcolor{DrawAIHeatGray!22}
& & \multicolumn{7}{c}{Fidelity $\Delta$} & \multicolumn{6}{c}{Editability $\Delta$} \\
\cmidrule(lr){3-9}\cmidrule(l){10-15}
\rowcolor{DrawAIHeatGray!12}
Model & Harness & Global & Text & Image & Formula & Shape & Connector & Table & Text & Image & Formula & Shape & Connector & Table \\
\midrule
\modelopenai{GPT-5.5} & Codex & \cellcolor{DrawAIHeatBlue!8}+0.9 & \cellcolor{DrawAIHeatBlue!9}+2.9 & \cellcolor{DrawAIHeatBlue!18}+15.6 & \cellcolor{DrawAIHeatBlue!9}+3.4 & \cellcolor{DrawAIHeatBlue!7}+0.6 & \cellcolor{DrawAIHeatBlue!9}+2.2 & \cellcolor{DrawAIHeatRed!8}-1.9 & \cellcolor{DrawAIHeatBlue!24}+24.8 & \cellcolor{DrawAIHeatBlue!19}+17.8 & \cellcolor{DrawAIHeatBlue!13}+9.4 & \cellcolor{DrawAIHeatBlue!9}+2.8 & \cellcolor{DrawAIHeatBlue!15}+11.3 & \cellcolor{DrawAIHeatBlue!8}+0.9 \\
\modelclaude{Claude Opus 4.8} & Claude Code & \cellcolor{DrawAIHeatBlue!8}+1.2 & \cellcolor{DrawAIHeatRed!7}-0.1 & \cellcolor{DrawAIHeatBlue!21}+20.5 & \cellcolor{DrawAIHeatRed!9}-2.2 & \cellcolor{DrawAIHeatRed!7}-0.5 & \cellcolor{DrawAIHeatBlue!9}+3.1 & \cellcolor{DrawAIHeatGray!24}0.0 & \cellcolor{DrawAIHeatBlue!26}+27.8 & \cellcolor{DrawAIHeatBlue!27}+29.1 & \cellcolor{DrawAIHeatBlue!7}+0.4 & \cellcolor{DrawAIHeatBlue!7}+0.6 & \cellcolor{DrawAIHeatBlue!9}+3.5 & \cellcolor{DrawAIHeatRed!20}-18.5 \\
\modelopenai{GPT-5.6 Sol} & Codex & \cellcolor{DrawAIHeatGray!24}0.0 & \cellcolor{DrawAIHeatBlue!8}+1.1 & \cellcolor{DrawAIHeatBlue!23}+24.0 & \cellcolor{DrawAIHeatRed!8}-1.8 & \cellcolor{DrawAIHeatRed!7}-0.1 & \cellcolor{DrawAIHeatBlue!9}+2.7 & \cellcolor{DrawAIHeatBlue!11}+5.6 & \cellcolor{DrawAIHeatBlue!28}+30.7 & \cellcolor{DrawAIHeatBlue!30}+33.8 & \cellcolor{DrawAIHeatRed!12}-7.3 & \cellcolor{DrawAIHeatGray!24}0.0 & \cellcolor{DrawAIHeatBlue!7}+0.7 & \cellcolor{DrawAIHeatRed!13}-9.3 \\
\modelopenai{GPT-5.4} & Codex & \cellcolor{DrawAIHeatBlue!9}+3.3 & \cellcolor{DrawAIHeatBlue!8}+1.7 & \cellcolor{DrawAIHeatBlue!18}+15.9 & \cellcolor{DrawAIHeatRed!10}-4.9 & \cellcolor{DrawAIHeatRed!8}-0.8 & \cellcolor{DrawAIHeatBlue!9}+3.2 & \cellcolor{DrawAIHeatBlue!10}+3.7 & \cellcolor{DrawAIHeatBlue!25}+26.1 & \cellcolor{DrawAIHeatBlue!17}+14.8 & \cellcolor{DrawAIHeatBlue!24}+24.8 & \cellcolor{DrawAIHeatBlue!10}+4.4 & \cellcolor{DrawAIHeatBlue!9}+2.9 & \cellcolor{DrawAIHeatBlue!18}+15.7 \\
\modelgemini{Gemini 3.5 Flash} & Antigravity & \cellcolor{DrawAIHeatBlue!8}+2.2 & \cellcolor{DrawAIHeatBlue!8}+1.2 & \cellcolor{DrawAIHeatBlue!20}+19.5 & \cellcolor{DrawAIHeatRed!15}-11.2 & \cellcolor{DrawAIHeatBlue!8}+1.3 & \cellcolor{DrawAIHeatBlue!8}+0.8 & \cellcolor{DrawAIHeatBlue!16}+13.0 & \cellcolor{DrawAIHeatBlue!27}+29.5 & \cellcolor{DrawAIHeatBlue!22}+21.7 & \cellcolor{DrawAIHeatBlue!18}+16.6 & \cellcolor{DrawAIHeatBlue!7}+0.5 & \cellcolor{DrawAIHeatBlue!13}+8.9 & \cellcolor{DrawAIHeatRed!16}-13.0 \\
\modelminimax{MiniMax M3} & Claude Code & \cellcolor{DrawAIHeatBlue!8}+1.5 & \cellcolor{DrawAIHeatBlue!10}+3.8 & \cellcolor{DrawAIHeatBlue!24}+25.4 & \cellcolor{DrawAIHeatRed!16}-12.6 & \cellcolor{DrawAIHeatRed!10}-4.0 & \cellcolor{DrawAIHeatRed!10}-4.9 & \cellcolor{DrawAIHeatRed!8}-1.9 & \cellcolor{DrawAIHeatBlue!41}+54.0 & \cellcolor{DrawAIHeatBlue!33}+37.7 & \cellcolor{DrawAIHeatBlue!23}+23.3 & \cellcolor{DrawAIHeatBlue!8}+1.2 & \cellcolor{DrawAIHeatBlue!13}+9.2 & \cellcolor{DrawAIHeatBlue!15}+11.1 \\
\modelgemini{Gemini 3.1 Pro} & Antigravity & \cellcolor{DrawAIHeatBlue!9}+3.6 & \cellcolor{DrawAIHeatRed!7}-0.3 & \cellcolor{DrawAIHeatBlue!21}+20.3 & \cellcolor{DrawAIHeatRed!17}-14.5 & \cellcolor{DrawAIHeatRed!12}-7.3 & \cellcolor{DrawAIHeatRed!14}-11.0 & \cellcolor{DrawAIHeatRed!11}-5.6 & \cellcolor{DrawAIHeatBlue!33}+38.9 & \cellcolor{DrawAIHeatBlue!22}+22.5 & \cellcolor{DrawAIHeatBlue!15}+11.1 & \cellcolor{DrawAIHeatBlue!8}+0.9 & \cellcolor{DrawAIHeatBlue!13}+8.2 & \cellcolor{DrawAIHeatBlue!10}+3.7 \\
e\modelmimo{MiMo V2.5 Pro} & Claude Code & \cellcolor{DrawAIHeatRed!8}-1.5 & \cellcolor{DrawAIHeatBlue!25}+26.5 & \cellcolor{DrawAIHeatBlue!24}+24.3 & \cellcolor{DrawAIHeatBlue!18}+15.6 & \cellcolor{DrawAIHeatRed!10}-4.0 & \cellcolor{DrawAIHeatBlue!7}+0.3 & \cellcolor{DrawAIHeatBlue!20}+18.5 & \cellcolor{DrawAIHeatBlue!27}+28.9 & \cellcolor{DrawAIHeatBlue!12}+8.1 & \cellcolor{DrawAIHeatBlue!41}+53.4 & \cellcolor{DrawAIHeatBlue!35}+40.5 & \cellcolor{DrawAIHeatBlue!41}+76.5 & \cellcolor{DrawAIHeatBlue!41}+63.9 \\
\modelmimo{MiMo V2.5} & Claude Code & \cellcolor{DrawAIHeatBlue!8}+1.8 & \cellcolor{DrawAIHeatBlue!9}+2.5 & \cellcolor{DrawAIHeatBlue!25}+26.7 & \cellcolor{DrawAIHeatRed!22}-21.5 & \cellcolor{DrawAIHeatRed!9}-3.5 & \cellcolor{DrawAIHeatRed!13}-8.4 & \cellcolor{DrawAIHeatRed!16}-13.0 & \cellcolor{DrawAIHeatBlue!13}+9.1 & \cellcolor{DrawAIHeatBlue!39}+46.6 & \cellcolor{DrawAIHeatBlue!41}+52.5 & \cellcolor{DrawAIHeatBlue!10}+3.7 & \cellcolor{DrawAIHeatBlue!11}+5.7 & \cellcolor{DrawAIHeatBlue!11}+5.6 \\
\bottomrule
\end{tabular}
\end{table*}

\paragraph{Mean gains coexist with instance-level risk.}
The 10th percentile of per-image gain is negative in twelve of fourteen settings, and the share of improved images ranges from 67.5--90.0\%. Image Fidelity and Text and Image Editability improve for every model, whereas Table Editability regresses for three strong models. DrawAI-Flow therefore concentrates its benefit on each model's unresolved structural weaknesses without guaranteeing an improvement on every artifact.

\subsection{DrawAI-Flow Component Ablation}
\label{sec:flow-component-ablation}

We isolate the contribution of the two perceptual tools used by the Parser Agent through four controlled \modelopenai{GPT-5.5}--Codex settings. \emph{Full DrawAI-Flow} denotes the complete \texttt{default\_drawai\_dag}; the other variants remove OCR evidence, SAM3 evidence, or both while keeping the agent, reconstruction stage, and evaluation protocol fixed. Every setting is evaluated on the same matched artifact panel with the complete 39-criterion protocol. As elsewhere in the paper, decreasing metrics are first converted to $1-s$, tertiary metrics are aggregated into asset categories, Fidelity and Editability are macro-averaged over their available categories, and Overall weights the two dimensions equally. Tables~\ref{tab:flow-component-ablation} and~\ref{tab:flow-component-mechanisms} report the aggregate, paired, mechanism-specific, and domain-level results.

\paragraph{Aggregate effects.}
The complete workflow obtains 92.82 Fidelity, 89.07 Editability, and 90.95 Overall. Removing OCR reduces Overall by 3.61 points, primarily through a 6.42-point Editability loss, whereas removing SAM3 reduces Overall by 3.27 points with a larger relative effect on Fidelity. The paired comparison confirms that these are not driven by a small number of artifacts: Full DrawAI-Flow scores higher on 77.5\% of artifacts without OCR, 67.5\% without SAM3, and 85.0\% when both tools are removed. Joint removal lowers Overall by 8.16 points and Editability by 12.81 points. Its penalty exceeds the sum of the two isolated penalties by 1.28 Overall points and 2.12 Editability points, indicating that the two evidence sources are complementary rather than interchangeable.

\begin{table*}[t]
\centering
\caption{\textbf{Controlled component ablation of DrawAI-Flow.} All settings use GPT-5.5 with Codex and the same matched artifact panel. Panel (a) reports absolute scores and differences from the full workflow. Panel (b) reports paired mean $\pm$ standard deviation; negative values indicate degradation, and Full better is the percentage of artifacts on which the full workflow obtains a higher Overall score.}
\label{tab:flow-component-ablation}
\renewcommand{\arraystretch}{1.08}
\setlength{\tabcolsep}{7pt}
\textbf{(a) Aggregate scores}\par\vspace{3pt}
\begin{tabular}{lccccc}
\toprule
\rowcolor{DrawAIHeatGray!22}
Setting & Fidelity & Editability & Overall & $\Delta$ Overall \\
\midrule
\textbf{Full DrawAI-Flow} & \cellcolor{DrawAIHeatBlue!34}\textbf{92.82} & \cellcolor{DrawAIHeatBlue!31}\textbf{89.07} & \cellcolor{DrawAIHeatBlue!33}\textbf{90.95} & \cellcolor{DrawAIHeatGray!12}\textbf{0.00} \\
w/o OCR & \cellcolor{DrawAIHeatBlue!33}92.02 & \cellcolor{DrawAIHeatBlue!28}82.65 & \cellcolor{DrawAIHeatBlue!30}87.34 & \cellcolor{DrawAIHeatRed!15}-3.61 \\
w/o SAM3 & \cellcolor{DrawAIHeatBlue!32}90.56 & \cellcolor{DrawAIHeatBlue!29}84.80 & \cellcolor{DrawAIHeatBlue!31}87.68 & \cellcolor{DrawAIHeatRed!14}-3.27 \\
w/o OCR \& SAM3 & \cellcolor{DrawAIHeatBlue!32}89.31 & \cellcolor{DrawAIHeatBlue!24}76.26 & \cellcolor{DrawAIHeatBlue!28}82.79 & \cellcolor{DrawAIHeatRed!23}-8.16 \\
\bottomrule
\end{tabular}
\par\vspace{8pt}
\textbf{(b) Paired changes relative to Full DrawAI-Flow}\par\vspace{3pt}
\begin{tabular}{lccccc}
\toprule
\rowcolor{DrawAIHeatGray!22}
Ablation & $\Delta$ Fidelity & $\Delta$ Editability & $\Delta$ Overall & Full better \\
\midrule
w/o OCR & \cellcolor{DrawAIHeatRed!10}$-0.80\!\pm\!4.73$ & \cellcolor{DrawAIHeatRed!20}$-6.42\!\pm\!12.26$ & \cellcolor{DrawAIHeatRed!15}$-3.61\!\pm\!7.46$ & \cellcolor{DrawAIHeatBlue!24}77.5\% \\
w/o SAM3 & \cellcolor{DrawAIHeatRed!12}$-2.27\!\pm\!5.71$ & \cellcolor{DrawAIHeatRed!16}$-4.27\!\pm\!12.23$ & \cellcolor{DrawAIHeatRed!14}$-3.27\!\pm\!7.75$ & \cellcolor{DrawAIHeatBlue!18}67.5\% \\
w/o OCR \& SAM3 & \cellcolor{DrawAIHeatRed!15}$-3.51\!\pm\!5.67$ & \cellcolor{DrawAIHeatRed!32}$-12.81\!\pm\!15.39$ & \cellcolor{DrawAIHeatRed!23}$-8.16\!\pm\!9.62$ & \cellcolor{DrawAIHeatBlue!29}85.0\% \\
\bottomrule
\end{tabular}
\end{table*}

\begin{table*}[t]
\centering
\caption{\textbf{Mechanism and domain breakdown of the DrawAI-Flow ablation.} Global Fidelity denotes the Global Visual Fidelity category. Darker blue indicates a higher score.}
\label{tab:flow-component-mechanisms}
\renewcommand{\arraystretch}{1.08}
\setlength{\tabcolsep}{6pt}
\textbf{(a) Mechanism-specific categories}\par\vspace{3pt}
\begin{tabular}{lccccc}
\toprule
\rowcolor{DrawAIHeatGray!22}
Setting & Text Fidelity & Text Editability & Image Fidelity & Image Editability & Global Fidelity \\
\midrule
\textbf{Full DrawAI-Flow} & \cellcolor{DrawAIHeatBlue!35}\textbf{95.73} & \cellcolor{DrawAIHeatBlue!26}\textbf{80.75} & \cellcolor{DrawAIHeatBlue!29}\textbf{85.81} & \cellcolor{DrawAIHeatBlue!30}\textbf{87.26} & \cellcolor{DrawAIHeatBlue!31}\textbf{88.25} \\
w/o OCR & \cellcolor{DrawAIHeatBlue!34}92.84 & \cellcolor{DrawAIHeatBlue!13}59.09 & \cellcolor{DrawAIHeatBlue!29}85.19 & \cellcolor{DrawAIHeatBlue!29}84.52 & \cellcolor{DrawAIHeatBlue!30}86.79 \\
w/o SAM3 & \cellcolor{DrawAIHeatBlue!35}95.41 & \cellcolor{DrawAIHeatBlue!24}76.94 & \cellcolor{DrawAIHeatBlue!23}74.35 & \cellcolor{DrawAIHeatBlue!20}70.30 & \cellcolor{DrawAIHeatBlue!29}85.80 \\
w/o OCR \& SAM3 & \cellcolor{DrawAIHeatBlue!34}92.79 & \cellcolor{DrawAIHeatBlue!10}52.84 & \cellcolor{DrawAIHeatBlue!20}69.61 & \cellcolor{DrawAIHeatBlue!14}60.29 & \cellcolor{DrawAIHeatBlue!30}86.32 \\
\bottomrule
\end{tabular}
\par\vspace{8pt}
\textbf{(b) Overall score by domain}\par\vspace{3pt}
\begin{tabular}{lccccc}
\toprule
\rowcolor{DrawAIHeatGray!22}
Setting & Diagrams & Posters & Presentation slides & Scientific figures \\
\midrule
\textbf{Full DrawAI-Flow} & \cellcolor{DrawAIHeatBlue!33}\textbf{91.29} & \cellcolor{DrawAIHeatBlue!30}\textbf{86.08} & \cellcolor{DrawAIHeatBlue!34}\textbf{93.18} & \cellcolor{DrawAIHeatBlue!34}\textbf{93.23} \\
w/o OCR & \cellcolor{DrawAIHeatBlue!31}89.01 & \cellcolor{DrawAIHeatBlue!26}80.62 & \cellcolor{DrawAIHeatBlue!31}88.84 & \cellcolor{DrawAIHeatBlue!33}90.87 \\
w/o SAM3 & \cellcolor{DrawAIHeatBlue!30}86.77 & \cellcolor{DrawAIHeatBlue!29}84.72 & \cellcolor{DrawAIHeatBlue!32}90.78 & \cellcolor{DrawAIHeatBlue!31}88.45 \\
w/o OCR \& SAM3 & \cellcolor{DrawAIHeatBlue!29}84.88 & \cellcolor{DrawAIHeatBlue!23}75.48 & \cellcolor{DrawAIHeatBlue!30}86.37 & \cellcolor{DrawAIHeatBlue!29}84.41 \\
\bottomrule
\end{tabular}
\end{table*}

\paragraph{Mechanism-specific effects.}
The targeted categories make the division of labor explicit. Without OCR, Text Editability falls from 80.75 to 59.09 ($-21.66$), while Text Fidelity falls by only 2.89 points. OCR therefore contributes chiefly by grounding recognized content as reusable text objects rather than merely preserving visible text. Without SAM3, Image Fidelity falls by 11.46 points and Image Editability by 16.96 points, consistent with SAM3 supplying region and object-boundary evidence needed for localized image assets. Removing both tools compounds these failures: Text Editability and Image Editability decline by 27.91 and 26.97 points, respectively. In contrast, Global Fidelity changes by fewer than 2.5 points in every ablation. Page-level appearance can therefore conceal severe losses in object structure, reinforcing the need for the asset-level Fidelity and Editability taxonomy.

\paragraph{Domain sensitivity.}
Every domain benefits from both signals, but their relative importance follows the dominant content. OCR has its largest isolated effect on Posters and Presentation slides, reducing Overall by 5.45 and 4.34 points, respectively. SAM3 has its largest effect on Scientific figures and Diagrams, with drops of 4.78 and 4.53 points. Joint removal is most damaging for Posters ($-10.60$) and Scientific figures ($-8.82$). Relative to the sum of the isolated losses, the joint penalty is 3.78 points larger for Posters and 1.68 points larger for Scientific figures, approximately additive for Presentation slides, and mildly sub-additive for Diagrams. These differences show that the Parser Agent benefits from both semantic text evidence and spatial region evidence, while the balance between them depends on the composition being reconstructed.

\subsection{Comparison with Specialized Projects}
\label{sec:project-comparison-results}

\begin{figure*}[tbp]
\centering
\includegraphics[width=0.4\textwidth]{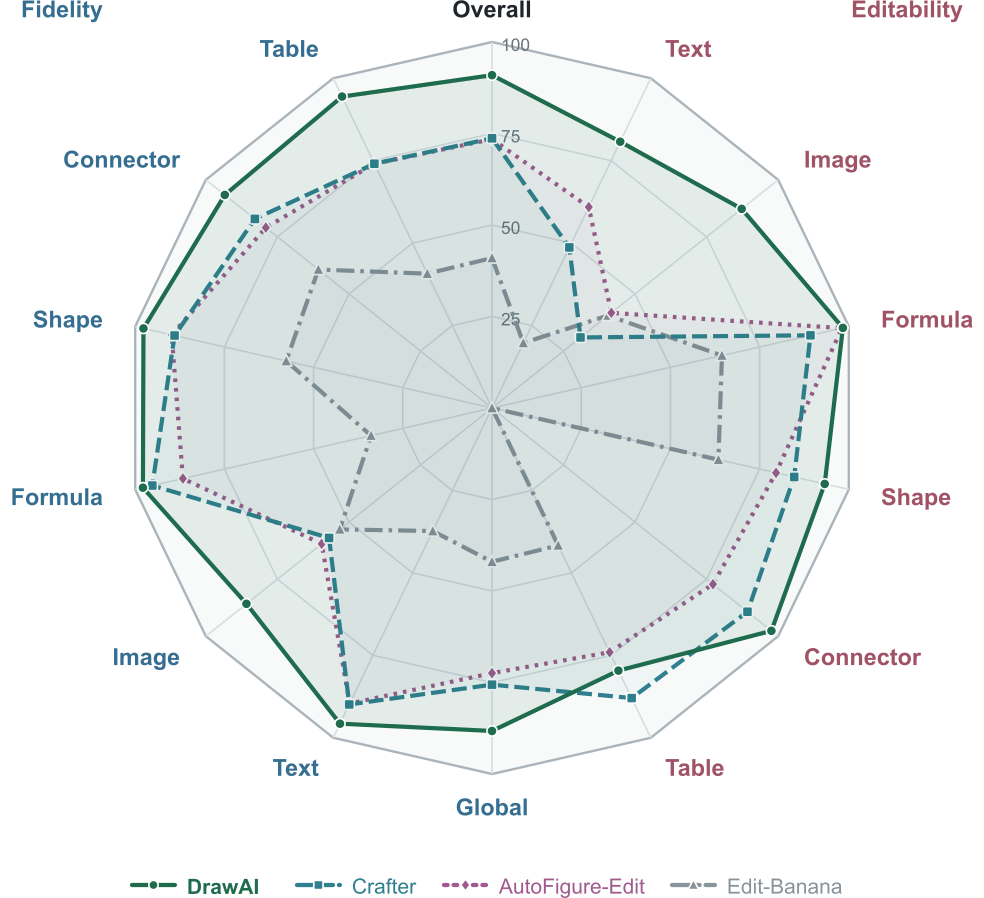}
\caption{\textbf{Comparison with image-to-editable artifact projects.} All methods are executed by \modelopenai{GPT-5.5}.}
\label{fig:project-comparison}
\end{figure*}

Using \modelopenai{GPT-5.5} on the same 80-image panel, DrawAI reaches 91.0 Overall (92.8 Fidelity and 89.1 Editability), compared with 73.7 for Crafter, 73.4 for AutoFigure-Edit, and 41.0 for Edit-Banana. Figure~\ref{fig:project-comparison} summarizes the comparison.

\paragraph{Specialized systems remain Fidelity-skewed.}
Fidelity exceeds Editability by 12.5 points for AutoFigure-Edit, 14.8 for Crafter, and 15.7 for Edit-Banana, versus 3.8 for DrawAI. The rule- and rubric-based branches also disagree about the runner-up: rubric-only scoring favors Crafter over AutoFigure-Edit (86.7 versus 81.0), whereas rule-only scoring favors AutoFigure-Edit over Crafter (69.9 versus 63.5). This disagreement motivates the hybrid protocol rather than either branch in isolation.

\paragraph{Criterion-level frontier.}
DrawAI obtains the best value on 35 of 39 criteria. The exceptions are formula-symbol preservation, where Crafter holds a near-tie, and the three table-structure criteria, where Crafter leads. DrawAI's largest margins concern editable image objects, image alignment, and visual hierarchy, but exact image placement and native table structure remain open problems. Tables are also the taxonomy slice with the lowest human agreement in Section~\ref{sec:human-validation}, making them difficult both to reconstruct and to judge. The former exhaustive 39-row project comparison table is omitted; Appendix~\ref{sec:complete-evaluation-criteria} retains the complete criterion definitions.

\subsection{Cost Analysis}
\label{sec:cost-analysis}

\begin{figure*}[t]
\centering
\includegraphics[width=0.5\textwidth]{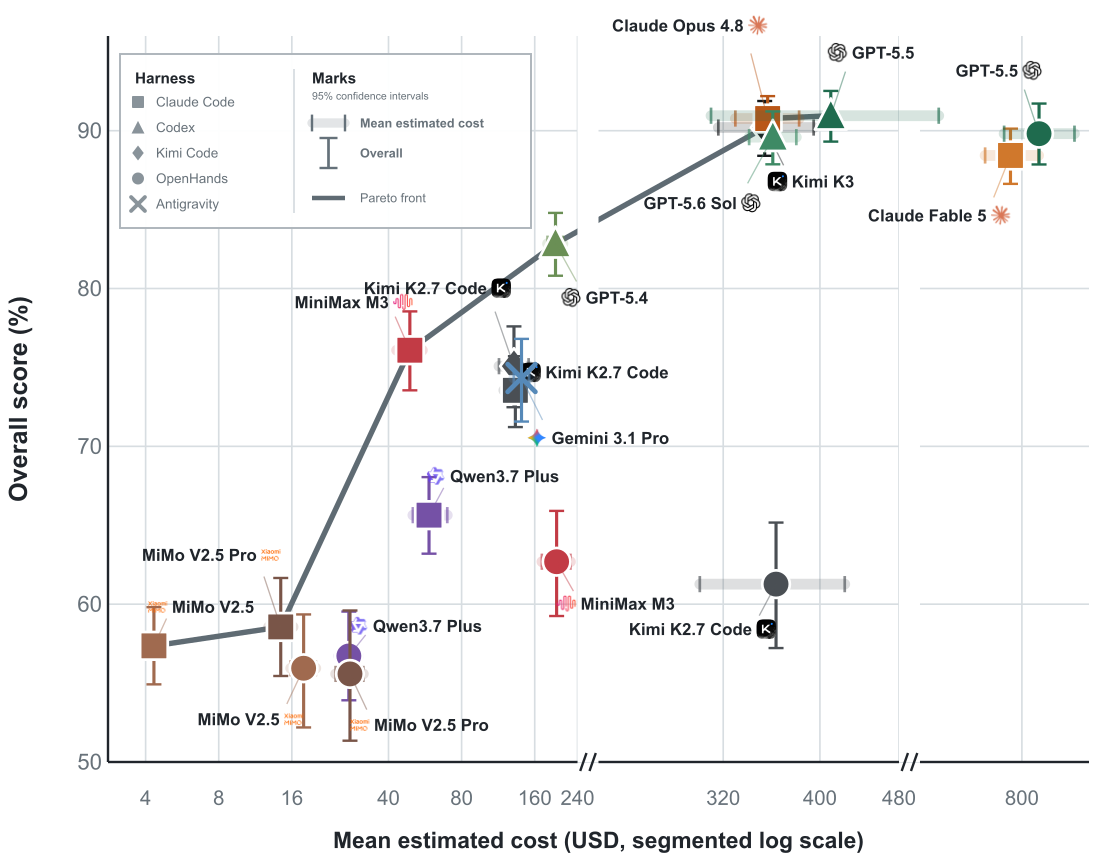}
\caption{\textbf{Mean estimated cost--quality Pareto frontier.} Each marker is a model--harness setting and reports its mean estimated API-equivalent cost. Bars show bootstrap 95\% confidence intervals. The line joins non-dominated settings, colors encode model families, and shapes encode harnesses.}
\label{fig:cost-utility}
\end{figure*}

We compare model--harness systems by their mean estimated API-equivalent cost for processing the complete 80-image DrawAI-Bench. The estimate is derived from actual request-level token records, separated into Input, Cached Input, and Output, using each provider's public standard rates, including cache-specific prices and context-length tiers; Qwen3.7 Plus Cached Input uses the official cache-hit price. We use \emph{estimated} because the experiments run through subscribed coding or token plans rather than direct metered API billing. Overall uses every evaluated artifact, whereas cost statistics use artifacts with complete token records.

Figure~\ref{fig:cost-utility} shows a broad frontier rather than one best operating point. MiniMax M3 and GPT-5.4 occupy intermediate cost--quality regions, while the strongest Claude, Kimi, and GPT configurations define a tightly grouped high-quality end. Several expensive settings are dominated on both axes, so model price alone does not predict artifact quality.

Harness choice moves both axes. For example, the mean estimated cost of running GPT-5.5 is more than twice as high with OpenHands as with Codex (approximately \$872 versus \$410), while its Overall score is lower (89.8 versus 91.0); related inefficiencies occur for Kimi K2.7 Code and MiniMax M3. Cost, like capability, is therefore a property of the complete execution configuration.

Table~\ref{tab:cost-quality-operating-points} reports the token use, estimated cost, uncertainty, and quality of the configurations underlying Figure~\ref{fig:cost-utility}. The nineteen settings span a $201\times$ cost range but only a 35.4-point Overall range.

\begin{table*}[t]
\centering
\caption{Resource, cost, and quality operating points underlying Figure~\ref{fig:cost-utility}, sorted by Overall in descending order. Token columns are logged means per artifact. Mean estimated cost and Overall are reported as mean $\pm$ sample standard deviation for the complete 80-artifact benchmark. Estimated cost denotes the API-equivalent value calculated from logged token use and public prices rather than realized API billing. Input combines uncached input and cache writes, while Cached Input is priced at each provider's cache-hit rate. Darker amber denotes greater resource use or estimated cost, and darker blue higher Overall quality.}
\label{tab:cost-quality-operating-points}
\setlength{\tabcolsep}{4.0pt}
\begin{tabular}{llrrrrr}
\toprule
\rowcolor{DrawAIHeatGray!22}
Model & Harness & Input & Cached Input & Output & Mean estimated cost (USD) & Overall (mean $\pm$ SD) \\
\midrule
\modelopenai{GPT-5.5} & Codex & \cellcolor{DrawAIHeatAmber!14}296k & \cellcolor{DrawAIHeatAmber!10}2.84M & \cellcolor{DrawAIHeatAmber!14}74k & \cellcolor{DrawAIHeatAmber!34}\$410.13 $\pm$ 495.42 & \cellcolor{DrawAIHeatBlue!28}\textbf{91.0 $\pm$ 7.4} \\
\modelclaude{Claude Opus 4.8} & Claude Code & \cellcolor{DrawAIHeatAmber!6}160k & \cellcolor{DrawAIHeatAmber!9}2.75M & \cellcolor{DrawAIHeatAmber!9}58k & \cellcolor{DrawAIHeatAmber!33}\$354.54 $\pm$ 120.40 & \cellcolor{DrawAIHeatBlue!28}\textbf{90.8 $\pm$ 6.8} \\
\modelkimi{Kimi K3} & Kimi Code & \cellcolor{DrawAIHeatAmber!14}295k & \cellcolor{DrawAIHeatAmber!20}6.39M & \cellcolor{DrawAIHeatAmber!20}107k & \cellcolor{DrawAIHeatAmber!33}\$352.14 $\pm$ 177.41 & \cellcolor{DrawAIHeatBlue!28}\textbf{90.2 $\pm$ 7.9} \\
\modelopenai{GPT-5.5} & OpenHands & \cellcolor{DrawAIHeatAmber!30}931k & \cellcolor{DrawAIHeatAmber!14}3.86M & \cellcolor{DrawAIHeatAmber!25}144k & \cellcolor{DrawAIHeatAmber!39}\$871.90 $\pm$ 707.98 & \cellcolor{DrawAIHeatBlue!27}\textbf{89.8 $\pm$ 8.9} \\
\modelopenai{GPT-5.6 Sol} & Codex & \cellcolor{DrawAIHeatAmber!12}249k & \cellcolor{DrawAIHeatAmber!9}2.77M & \cellcolor{DrawAIHeatAmber!10}62k & \cellcolor{DrawAIHeatAmber!33}\$358.79 $\pm$ 88.84 & \cellcolor{DrawAIHeatBlue!27}\textbf{89.6 $\pm$ 7.8} \\
\modelclaude{Claude Fable 5} & Claude Code & \cellcolor{DrawAIHeatAmber!8}189k & \cellcolor{DrawAIHeatAmber!8}2.44M & \cellcolor{DrawAIHeatAmber!11}66k & \cellcolor{DrawAIHeatAmber!38}\$755.66 $\pm$ 469.70 & \cellcolor{DrawAIHeatBlue!27}\textbf{88.4 $\pm$ 7.9} \\
\modelopenai{GPT-5.4} & Codex & \cellcolor{DrawAIHeatAmber!14}297k & \cellcolor{DrawAIHeatAmber!6}2.12M & \cellcolor{DrawAIHeatAmber!14}78k & \cellcolor{DrawAIHeatAmber!30}\$194.86 $\pm$ 64.29 & \cellcolor{DrawAIHeatBlue!26}\textbf{82.8 $\pm$ 9.0} \\
\modelminimax{MiniMax M3} & Claude Code & \cellcolor{DrawAIHeatAmber!23}565k & \cellcolor{DrawAIHeatAmber!19}5.90M & \cellcolor{DrawAIHeatAmber!13}74k & \cellcolor{DrawAIHeatAmber!21}\$48.98 $\pm$ 27.86 & \cellcolor{DrawAIHeatBlue!24}\textbf{76.1 $\pm$ 11.3} \\
\modelkimi{Kimi K2.7 Code} & Kimi Code & \cellcolor{DrawAIHeatAmber!12}245k & \cellcolor{DrawAIHeatAmber!19}5.71M & \cellcolor{DrawAIHeatAmber!15}81k & \cellcolor{DrawAIHeatAmber!27}\$131.40 $\pm$ 74.85 & \cellcolor{DrawAIHeatBlue!24}\textbf{75.1 $\pm$ 11.8} \\
\modelgemini{Gemini 3.1 Pro} & Antigravity & \cellcolor{DrawAIHeatAmber!14}294k & \cellcolor{DrawAIHeatAmber!9}2.59M & \cellcolor{DrawAIHeatAmber!8}55k & \cellcolor{DrawAIHeatAmber!28}\$141.16 $\pm$ 46.74 & \cellcolor{DrawAIHeatBlue!24}\textbf{74.3 $\pm$ 12.1} \\
\modelkimi{Kimi K2.7 Code} & Claude Code & \cellcolor{DrawAIHeatAmber!12}241k & \cellcolor{DrawAIHeatAmber!19}6.03M & \cellcolor{DrawAIHeatAmber!13}72k & \cellcolor{DrawAIHeatAmber!27}\$133.15 $\pm$ 64.70 & \cellcolor{DrawAIHeatBlue!23}\textbf{73.5 $\pm$ 10.3} \\
\modelqwen{Qwen3.7 Plus} & Claude Code & \cellcolor{DrawAIHeatAmber!39}1.84M & \cellcolor{DrawAIHeatAmber!13}3.74M & \cellcolor{DrawAIHeatAmber!16}84k & \cellcolor{DrawAIHeatAmber!22}\$58.74 $\pm$ 45.74 & \cellcolor{DrawAIHeatBlue!21}\textbf{65.6 $\pm$ 11.3} \\
\modelminimax{MiniMax M3} & OpenHands & \cellcolor{DrawAIHeatAmber!35}1.35M & \cellcolor{DrawAIHeatAmber!39}28.11M & \cellcolor{DrawAIHeatAmber!39}307k & \cellcolor{DrawAIHeatAmber!30}\$196.88 $\pm$ 121.53 & \cellcolor{DrawAIHeatBlue!21}\textbf{62.7 $\pm$ 15.6} \\
\modelkimi{Kimi K2.7 Code} & OpenHands & \cellcolor{DrawAIHeatAmber!28}800k & \cellcolor{DrawAIHeatAmber!31}15.27M & \cellcolor{DrawAIHeatAmber!33}214k & \cellcolor{DrawAIHeatAmber!34}\$361.46 $\pm$ 237.75 & \cellcolor{DrawAIHeatBlue!20}\textbf{61.3 $\pm$ 15.9} \\
\modelmimo{MiMo V2.5 Pro} & Claude Code & \cellcolor{DrawAIHeatAmber!12}248k & \cellcolor{DrawAIHeatAmber!19}5.92M & \cellcolor{DrawAIHeatAmber!9}59k & \cellcolor{DrawAIHeatAmber!13}\$14.40 $\pm$ 7.61 & \cellcolor{DrawAIHeatBlue!20}\textbf{58.6 $\pm$ 14.1} \\
\modelmimo{MiMo V2.5} & Claude Code & \cellcolor{DrawAIHeatAmber!9}207k & \cellcolor{DrawAIHeatAmber!14}4.11M & \cellcolor{DrawAIHeatAmber!6}49k & \cellcolor{DrawAIHeatAmber!6}\$4.33 $\pm$ 1.66 & \cellcolor{DrawAIHeatBlue!19}\textbf{57.4 $\pm$ 11.3} \\
\modelqwen{Qwen3.7 Plus} & OpenHands & \cellcolor{DrawAIHeatAmber!22}523k & \cellcolor{DrawAIHeatAmber!14}4.08M & \cellcolor{DrawAIHeatAmber!15}78k & \cellcolor{DrawAIHeatAmber!17}\$27.46 $\pm$ 11.76 & \cellcolor{DrawAIHeatBlue!19}\textbf{56.7 $\pm$ 12.9} \\
\modelmimo{MiMo V2.5} & OpenHands & \cellcolor{DrawAIHeatAmber!29}908k & \cellcolor{DrawAIHeatAmber!34}18.34M & \cellcolor{DrawAIHeatAmber!27}161k & \cellcolor{DrawAIHeatAmber!15}\$17.89 $\pm$ 9.85 & \cellcolor{DrawAIHeatBlue!19}\textbf{55.9 $\pm$ 16.6} \\
\modelmimo{MiMo V2.5 Pro} & OpenHands & \cellcolor{DrawAIHeatAmber!22}505k & \cellcolor{DrawAIHeatAmber!26}10.41M & \cellcolor{DrawAIHeatAmber!19}104k & \cellcolor{DrawAIHeatAmber!18}\$27.76 $\pm$ 16.95 & \cellcolor{DrawAIHeatBlue!19}\textbf{55.6 $\pm$ 18.9} \\
\bottomrule
\end{tabular}
\end{table*}

\paragraph{Marginal cost varies by two orders of magnitude.}
Seven settings are Pareto-optimal. Moving from \modelmimo{MiMo V2.5 Pro} to \modelminimax{MiniMax M3} costs roughly \$2 per additional Overall point, mid-frontier steps cost about \$21--22 per point, and the last 0.2-point step from \modelclaude{Claude Opus 4.8} to \modelopenai{GPT-5.5} costs roughly \$278 per point. Several expensive settings are strictly dominated on both axes.

\paragraph{Harness costs reflect execution trajectories.}
For five of six models with both OpenHands and an alternative harness, OpenHands is 1.9--4.1$\times$ more expensive at equal or lower quality because it produces longer cached-input and output trajectories. \modelqwen{Qwen3.7 Plus} is the exception: Claude Code costs more but returns 8.9 additional Overall points. Cost must therefore be attributed to the complete model--harness configuration.

\paragraph{Price, trajectory length, and consistency are distinct.}
The Claude settings use compact trajectories but remain expensive because of provider unit prices, while MiMo settings show the reverse pattern. Quality and consistency move together ($r=-0.89$ between mean Overall and its per-artifact standard deviation). Per-artifact costs are also heavy-tailed: \modelopenai{GPT-5.5} under Codex is the only setting whose cost standard deviation exceeds its mean, so mean cost alone understates budget risk.

\section{Conclusion}
\label{sec:conclusion}

To bridge the gap between high-quality visual synthesis and practical authoring, we formulate and systematically investigate the task of image-to-editable reconstruction. To rigorously evaluate this capability, we introduce \textbf{DrawAI-Bench}, a cross-domain framework designed to jointly measure visual Fidelity and practical Editability. And we further develop \textbf{DrawAI-Flow}, a robust agentic workflow driven by explicit reconstruction planning and iterative image-as-code synthesis.
Our systematic evaluation reveals three consistent insights: leading models achieve similar overall performance despite exhibiting distinctly different asset-level capabilities, the choice of agentic harness materially dictates success, and structured procedural support is the primary driver for recovering genuinely manipulable structures. 
Ultimately, DrawAI maps both the promise and the current boundaries of multimodal agents. It demonstrates that reliable image-to-editable conversion is not an out-of-the-box capability achievable through direct prompting alone, but rather an emergent property of a fully integrated agentic ecosystem.

\bibliographystyle{unsrtnat}
\bibliography{references}

\begin{thebibliography}{49}
\providecommand{\natexlab}[1]{#1}
\providecommand{\url}[1]{\texttt{#1}}
\expandafter\ifx\csname urlstyle\endcsname\relax
  \providecommand{\doi}[1]{doi: #1}\else
  \providecommand{\doi}{doi: \begingroup \urlstyle{rm}\Url}\fi

\bibitem[Ramesh et~al.(2021)Ramesh, Pavlov, Goh, Gray, Voss, Radford, Chen, and
  Sutskever]{ramesh2021zeroshot}
Aditya Ramesh, Mikhail Pavlov, Gabriel Goh, Scott Gray, Chelsea Voss, Alec
  Radford, Mark Chen, and Ilya Sutskever.
\newblock Zero-shot text-to-image generation.
\newblock In \emph{Proceedings of the 38th International Conference on Machine
  Learning}, volume 139 of \emph{Proceedings of Machine Learning Research},
  pages 8821--8831. PMLR, 2021.
\newblock URL \url{https://proceedings.mlr.press/v139/ramesh21a.html}.

\bibitem[Nichol et~al.(2022)Nichol, Dhariwal, Ramesh, Shyam, Mishkin, McGrew,
  Sutskever, and Chen]{nichol2021glide}
Alexander~Quinn Nichol, Prafulla Dhariwal, Aditya Ramesh, Pranav Shyam, Pamela
  Mishkin, Bob McGrew, Ilya Sutskever, and Mark Chen.
\newblock {GLIDE}: Towards photorealistic image generation and editing with
  text-guided diffusion models.
\newblock In \emph{Proceedings of the 39th International Conference on Machine
  Learning}, volume 162 of \emph{Proceedings of Machine Learning Research},
  pages 16784--16804. PMLR, 2022.
\newblock URL \url{https://proceedings.mlr.press/v162/nichol22a.html}.

\bibitem[Saharia et~al.(2022)Saharia, Chan, Saxena, Li, Whang, Denton,
  Ghasemipour, Gontijo~Lopes, Karagol~Ayan, Salimans, Ho, Fleet, and
  Norouzi]{saharia2022imagen}
Chitwan Saharia, William Chan, Saurabh Saxena, Lala Li, Jay Whang, Emily~L.
  Denton, Kamyar Ghasemipour, Raphael Gontijo~Lopes, Burcu Karagol~Ayan, Tim
  Salimans, Jonathan Ho, David~J. Fleet, and Mohammad Norouzi.
\newblock Photorealistic text-to-image diffusion models with deep language
  understanding.
\newblock In \emph{Advances in Neural Information Processing Systems},
  volume~35, pages 36479--36494. Curran Associates, Inc., 2022.
\newblock \doi{10.52202/068431-2643}.
\newblock URL
  \url{https://proceedings.neurips.cc/paper_files/paper/2022/hash/ec795aeadae0b7d230fa35cbaf04c041-Abstract-Conference.html}.

\bibitem[Rombach et~al.(2022)Rombach, Blattmann, Lorenz, Esser, and
  Ommer]{rombach2022latent}
Robin Rombach, Andreas Blattmann, Dominik Lorenz, Patrick Esser, and Bj{\"o}rn
  Ommer.
\newblock High-resolution image synthesis with latent diffusion models.
\newblock In \emph{Proceedings of the IEEE/CVF Conference on Computer Vision
  and Pattern Recognition}, pages 10684--10695, 2022.
\newblock \doi{10.1109/CVPR52688.2022.01042}.
\newblock URL
  \url{https://openaccess.thecvf.com/content/CVPR2022/html/Rombach_High-Resolution_Image_Synthesis_With_Latent_Diffusion_Models_CVPR_2022_paper.html}.

\bibitem[Zhu et~al.(2026{\natexlab{a}})Zhu, Meng, Song, Wei, Li, Pfister, and
  Yoon]{zhu2026paperbanana}
Dawei Zhu, Rui Meng, Yale Song, Xiyu Wei, Sujian Li, Tomas Pfister, and Jinsung
  Yoon.
\newblock Paperbanana: Automating academic illustration for ai scientists,
  2026{\natexlab{a}}.
\newblock URL \url{https://arxiv.org/abs/2601.23265}.

\bibitem[Zhu et~al.(2026{\natexlab{b}})Zhu, Lin, Weng, Lu, Xie, Wei, Liu, Sun,
  and Zhang]{zhu2026autofigure}
Minjun Zhu, Zhen Lin, Yixuan Weng, Panzhong Lu, Qiujie Xie, Yifan Wei, Sifan
  Liu, Qiyao Sun, and Yue Zhang.
\newblock Autofigure: Generating and refining publication-ready scientific
  illustrations.
\newblock In \emph{The Fourteenth International Conference on Learning
  Representations}, 2026{\natexlab{b}}.
\newblock URL \url{https://openreview.net/forum?id=5N3z9JQJKq}.

\bibitem[Yao et~al.(2023)Yao, Zhao, Yu, Du, Shafran, Narasimhan, and
  Cao]{yao2023react}
Shunyu Yao, Jeffrey Zhao, Dian Yu, Nan Du, Izhak Shafran, Karthik Narasimhan,
  and Yuan Cao.
\newblock {ReAct}: Synergizing reasoning and acting in language models.
\newblock In \emph{International Conference on Learning Representations}, 2023.
\newblock URL \url{https://arxiv.org/abs/2210.03629}.

\bibitem[Yang et~al.(2023)Yang, Li, Wang, Lin, Azarnasab, Ahmed, Liu, Liu,
  Zeng, and Wang]{yang2023mmreact}
Zhengyuan Yang, Linjie Li, Jianfeng Wang, Kevin Lin, Ehsan Azarnasab, Faisal
  Ahmed, Zicheng Liu, Ce~Liu, Michael Zeng, and Lijuan Wang.
\newblock {MM-REACT}: Prompting {ChatGPT} for multimodal reasoning and action,
  2023.
\newblock URL \url{https://arxiv.org/abs/2303.11381}.

\bibitem[Wang et~al.(2024{\natexlab{a}})Wang, Li, Song, Xu, Tang, Zhuge, Pan,
  et~al.]{wang2024openhands}
Xingyao Wang, Boxuan Li, Yufan Song, Frank~F. Xu, Xiangru Tang, Mingchen Zhuge,
  Jiayi Pan, et~al.
\newblock Openhands: An open platform for ai software developers as generalist
  agents, 2024{\natexlab{a}}.
\newblock URL \url{https://arxiv.org/abs/2407.16741}.

\bibitem[Wang et~al.(2024{\natexlab{b}})Wang, Chen, Yuan, Zhang, Li, Peng, and
  Ji]{wang2024codeact}
Xingyao Wang, Yangyi Chen, Lifan Yuan, Yizhe Zhang, Yunzhu Li, Hao Peng, and
  Heng Ji.
\newblock Executable code actions elicit better {LLM} agents.
\newblock In \emph{Proceedings of the 41st International Conference on Machine
  Learning}, volume 235 of \emph{Proceedings of Machine Learning Research},
  pages 50208--50232. PMLR, 2024{\natexlab{b}}.
\newblock URL \url{https://proceedings.mlr.press/v235/wang24h.html}.

\bibitem[Rodriguez et~al.(2025)Rodriguez, Puri, Agarwal, Laradji, Rodriguez,
  Rajeswar, Vazquez, Pal, and Pedersoli]{rodriguez2025starvector}
Juan~A. Rodriguez, Abhay Puri, Shubham Agarwal, Issam~H. Laradji, Pau
  Rodriguez, Sai Rajeswar, David Vazquez, Christopher Pal, and Marco Pedersoli.
\newblock Starvector: Generating scalable vector graphics code from images and
  text.
\newblock In \emph{Proceedings of the IEEE/CVF Conference on Computer Vision
  and Pattern Recognition}, pages 16175--16186, 2025.
\newblock \doi{10.1109/CVPR52734.2025.01508}.
\newblock URL
  \url{https://openaccess.thecvf.com/content/CVPR2025/html/Rodriguez_StarVector_Generating_Scalable_Vector_Graphics_Code_from_Images_and_Text_CVPR_2025_paper.html}.

\bibitem[Esser et~al.(2024)Esser, Kulal, Blattmann, Entezari, M{\"u}ller,
  Saini, Levi, Lorenz, Sauer, Boesel, Podell, Dockhorn, English, and
  Rombach]{esser2024rectified}
Patrick Esser, Sumith Kulal, Andreas Blattmann, Rahim Entezari, Jonas
  M{\"u}ller, Harry Saini, Yam Levi, Dominik Lorenz, Axel Sauer, Frederic
  Boesel, Dustin Podell, Tim Dockhorn, Zion English, and Robin Rombach.
\newblock Scaling rectified flow transformers for high-resolution image
  synthesis.
\newblock In \emph{Proceedings of the 41st International Conference on Machine
  Learning}, volume 235 of \emph{Proceedings of Machine Learning Research},
  pages 12606--12633. PMLR, 2024.
\newblock URL \url{https://proceedings.mlr.press/v235/esser24a.html}.

\bibitem[Wu et~al.(2025)Wu, Li, Zhou, Lin, Gao, Yan, Yin, Bai, Xu, Chen, Chen,
  Tang, Zhang, Wang, Yang, Yu, Cheng, Liu, Li, Zhang, Meng, Wei, Ni, Chen, Cao,
  Peng, Qu, Wu, Wang, Yu, Wen, Feng, Xu, Wang, Zhang, Zhu, Wu, Cai, and
  Liu]{wu2025qwenimage}
Chenfei Wu, Jiahao Li, Jingren Zhou, Junyang Lin, Kaiyuan Gao, Kun Yan,
  Sheng-ming Yin, Shuai Bai, Xiao Xu, Yilei Chen, Yuxiang Chen, Zecheng Tang,
  Zekai Zhang, Zhengyi Wang, An~Yang, Bowen Yu, Chen Cheng, Dayiheng Liu,
  Deqing Li, Hang Zhang, Hao Meng, Hu~Wei, Jingyuan Ni, Kai Chen, Kuan Cao,
  Liang Peng, Lin Qu, Minggang Wu, Peng Wang, Shuting Yu, Tingkun Wen, Wensen
  Feng, Xiaoxiao Xu, Yi~Wang, Yichang Zhang, Yongqiang Zhu, Yujia Wu, Yuxuan
  Cai, and Zenan Liu.
\newblock Qwen-image technical report, 2025.
\newblock URL \url{https://arxiv.org/abs/2508.02324}.

\bibitem[Li et~al.(2023)Li, Liu, Wu, Mu, Yang, Gao, Li, and Lee]{li2023gligen}
Yuheng Li, Haotian Liu, Qingyang Wu, Fangzhou Mu, Jianwei Yang, Jianfeng Gao,
  Chunyuan Li, and Yong~Jae Lee.
\newblock Gligen: Open-set grounded text-to-image generation.
\newblock In \emph{Proceedings of the IEEE/CVF Conference on Computer Vision
  and Pattern Recognition}, pages 22511--22521, 2023.
\newblock \doi{10.1109/CVPR52729.2023.02156}.
\newblock URL
  \url{https://openaccess.thecvf.com/content/CVPR2023/html/Li_GLIGEN_Open-Set_Grounded_Text-to-Image_Generation_CVPR_2023_paper.html}.

\bibitem[Zhang et~al.(2023)Zhang, Rao, and Agrawala]{zhang2023controlnet}
Lvmin Zhang, Anyi Rao, and Maneesh Agrawala.
\newblock Adding conditional control to text-to-image diffusion models.
\newblock In \emph{Proceedings of the IEEE/CVF International Conference on
  Computer Vision}, pages 3836--3847, 2023.
\newblock URL
  \url{https://openaccess.thecvf.com/content/ICCV2023/html/Zhang_Adding_Conditional_Control_to_Text-to-Image_Diffusion_Models_ICCV_2023_paper.html}.

\bibitem[Mou et~al.(2024)Mou, Wang, Xie, Wu, Zhang, Qi, and
  Shan]{mou2024t2iadapter}
Chong Mou, Xintao Wang, Liangbin Xie, Yanze Wu, Jian Zhang, Zhongang Qi, and
  Ying Shan.
\newblock T2i-adapter: Learning adapters to dig out more controllable ability
  for text-to-image diffusion models.
\newblock \emph{Proceedings of the AAAI Conference on Artificial Intelligence},
  38\penalty0 (5):\penalty0 4296--4304, 2024.
\newblock \doi{10.1609/aaai.v38i5.28226}.
\newblock URL \url{https://ojs.aaai.org/index.php/AAAI/article/view/28226}.

\bibitem[Ye et~al.(2023)Ye, Zhang, Liu, Han, and Yang]{ye2023ipadapter}
Hu~Ye, Jun Zhang, Sibo Liu, Xiao Han, and Wei Yang.
\newblock Ip-adapter: Text compatible image prompt adapter for text-to-image
  diffusion models, 2023.
\newblock URL \url{https://arxiv.org/abs/2308.06721}.

\bibitem[Cao et~al.(2026)Cao, Zhou, Song, and Yang]{cao2025controllable}
Pu~Cao, Feng Zhou, Qing Song, and Lu~Yang.
\newblock Controllable generation with text-to-image diffusion models: A
  survey.
\newblock \emph{IEEE Transactions on Pattern Analysis and Machine
  Intelligence}, 48\penalty0 (4):\penalty0 4771--4791, 2026.
\newblock \doi{10.1109/TPAMI.2025.3646548}.
\newblock URL \url{https://doi.org/10.1109/TPAMI.2025.3646548}.

\bibitem[Cao et~al.(2025)Cao, Zhou, Yang, Huang, and Song]{cao2025imageallneed}
Pu~Cao, Feng Zhou, Lu~Yang, Tianrui Huang, and Qing Song.
\newblock Image is all you need to empower large-scale diffusion models for
  in-domain generation.
\newblock In \emph{Proceedings of the IEEE/CVF Conference on Computer Vision
  and Pattern Recognition}, pages 18358--18368, 2025.
\newblock \doi{10.1109/CVPR52734.2025.01711}.
\newblock URL
  \url{https://openaccess.thecvf.com/content/CVPR2025/html/Cao_Image_is_All_You_Need_to_Empower_Large-scale_Diffusion_Models_CVPR_2025_paper.html}.

\bibitem[Rodriguez et~al.(2023)Rodriguez, Vazquez, Laradji, Pedersoli, and
  Rodriguez]{rodriguez2023figgen}
Juan~A. Rodriguez, David Vazquez, Issam Laradji, Marco Pedersoli, and Pau
  Rodriguez.
\newblock Figgen: Text to scientific figure generation, 2023.
\newblock URL \url{https://arxiv.org/abs/2306.00800}.

\bibitem[Lin et~al.(2026)Lin, Xie, Zhu, Li, Sun, Gu, Ding, Sun, Guo, Lu, Ning,
  Weng, and Zhang]{lin2026autofigureedit}
Zhen Lin, Qiujie Xie, Minjun Zhu, Shichen Li, QiYao Sun, Enhao Gu, Yiran Ding,
  Ke~Sun, Fang Guo, Panzhong Lu, Zhiyuan Ning, Yixuan Weng, and Yue Zhang.
\newblock Autofigure-edit: Generating editable scientific illustrations via
  reference-guided styling.
\newblock In \emph{Proceedings of the 64th Annual Meeting of the Association
  for Computational Linguistics (Volume 3: System Demonstrations)}, pages
  57--67. Association for Computational Linguistics, 2026.
\newblock \doi{10.18653/v1/2026.acl-demo.6}.
\newblock URL \url{https://aclanthology.org/2026.acl-demo.6/}.

\bibitem[Zhao et~al.(2026)Zhao, Si, Wang, Wang, Chen, Li, Liang, Sun, and
  Zhang]{zhao2026crafter}
Haozhe Zhao, Shuzheng Si, Zhenhailong Wang, Zheng Wang, Liang Chen, Xiaotong
  Li, Zhixiang Liang, Maosong Sun, and Minjia Zhang.
\newblock Crafter: A multi-agent harness for editable scientific figure
  generation from diverse inputs, 2026.
\newblock URL \url{https://arxiv.org/abs/2605.30611}.

\bibitem[Sun et~al.(2025)Sun, Zhang, Feng, Li, Li, Ai, Chang, Dai, and
  Zhang]{sun2025pixels}
Jianwen Sun, Fanrui Zhang, Yukang Feng, Chuanhao Li, Zizhen Li, Jiaxin Ai,
  Yifan Chang, Yu~Dai, and Kaipeng Zhang.
\newblock From pixels to paths: A multi-agent framework for editable scientific
  illustration, 2025.
\newblock URL \url{https://arxiv.org/abs/2510.27452}.

\bibitem[Fu et~al.(2022)Fu, Wang, McDuff, and Song]{fu2022doc2ppt}
Tsu-Jui Fu, William~Yang Wang, Daniel McDuff, and Yale Song.
\newblock Doc2ppt: Automatic presentation slides generation from scientific
  documents.
\newblock \emph{Proceedings of the AAAI Conference on Artificial Intelligence},
  36\penalty0 (1):\penalty0 634--642, 2022.
\newblock \doi{10.1609/aaai.v36i1.19943}.
\newblock URL \url{https://ojs.aaai.org/index.php/AAAI/article/view/19943}.

\bibitem[Zheng et~al.(2025)Zheng, Guan, Kong, Zhang, Zheng, Zhou, Lin, Lu, Han,
  and Sun]{zheng2025pptagent}
Hao Zheng, Xinyan Guan, Hao Kong, Wenkai Zhang, Jia Zheng, Weixiang Zhou,
  Hongyu Lin, Yaojie Lu, Xianpei Han, and Le~Sun.
\newblock Pptagent: Generating and evaluating presentations beyond
  text-to-slides.
\newblock In \emph{Proceedings of the 2025 Conference on Empirical Methods in
  Natural Language Processing}, pages 14402--14418. Association for
  Computational Linguistics, 2025.
\newblock \doi{10.18653/v1/2025.emnlp-main.728}.
\newblock URL \url{https://aclanthology.org/2025.emnlp-main.728/}.

\bibitem[Pang et~al.(2025)Pang, Lin, Jian, He, and Torr]{pang2025paper2poster}
Wei Pang, Kevin~Qinghong Lin, Xiangru Jian, Xi~He, and Philip Torr.
\newblock Paper2poster: Towards multimodal poster automation from scientific
  papers.
\newblock In \emph{Advances in Neural Information Processing Systems}, 2025.
\newblock URL \url{https://arxiv.org/abs/2505.21497}.

\bibitem[Yang et~al.(2026)Yang, Li, Ren, Lu, Wang, Huang, Zong, Zhan, and
  Li]{yang2026slidesgenbench}
Yunqiao Yang, Wenbo Li, Houxing Ren, Zimu Lu, Ke~Wang, Zhiyuan Huang, Zhuofan
  Zong, Mingjie Zhan, and Hongsheng Li.
\newblock Slidesgen-bench: Evaluating slides generation via computational and
  quantitative metrics, 2026.
\newblock URL \url{https://arxiv.org/abs/2601.09487}.

\bibitem[Chen et~al.(2026)Chen, Zhu, Li, Wang, Yang, and
  Guo]{chen2026presentbench}
Xin-Sheng Chen, Jiayu Zhu, Pei-lin Li, Hanzheng Wang, Shuojin Yang, and
  Meng-Hao Guo.
\newblock Presentbench: A fine-grained rubric-based benchmark for slide
  generation, 2026.
\newblock URL \url{https://arxiv.org/abs/2603.07244}.

\bibitem[Jang et~al.(2026)Jang, Heisler, Xing, Li, Wang, Xiong, Zhang, and
  Fan]{jang2026deckbench}
Daesik Jang, Morgan~Lindsay Heisler, Linzi Xing, Yifei Li, Edward Wang, Ying
  Xiong, Yong Zhang, and Zhenan Fan.
\newblock Deckbench: Benchmarking multi-agent frameworks for academic slide
  generation and editing, 2026.
\newblock URL \url{https://arxiv.org/abs/2602.13318}.

\bibitem[Deng et~al.(2017)Deng, Kanervisto, Ling, and Rush]{deng2017markup}
Yuntian Deng, Anssi Kanervisto, Jeffrey Ling, and Alexander~M. Rush.
\newblock Image-to-markup generation with coarse-to-fine attention.
\newblock In \emph{Proceedings of the 34th International Conference on Machine
  Learning}, volume~70 of \emph{Proceedings of Machine Learning Research},
  pages 980--989. PMLR, 2017.
\newblock URL \url{https://proceedings.mlr.press/v70/deng17a.html}.

\bibitem[Yang et~al.(2025{\natexlab{a}})Yang, Shi, Liu, Shui, Wang, Jing, Xu,
  Zhu, Li, Zhang, Liu, Nie, Cai, and Yang]{yang2025chartmimic}
Cheng Yang, Chufan Shi, Yaxin Liu, Bo~Shui, Junjie Wang, Mohan Jing, Linran Xu,
  Xinyu Zhu, Siheng Li, Yuxiang Zhang, Gongye Liu, Xiaomei Nie, Deng Cai, and
  Yujiu Yang.
\newblock Chartmimic: Evaluating {LMM}'s cross-modal reasoning capability via
  chart-to-code generation.
\newblock In \emph{International Conference on Learning Representations},
  2025{\natexlab{a}}.
\newblock URL \url{https://arxiv.org/abs/2406.09961}.

\bibitem[Beltramelli(2017)]{beltramelli2017pix2code}
Tony Beltramelli.
\newblock pix2code: Generating code from a graphical user interface screenshot,
  2017.
\newblock URL \url{https://arxiv.org/abs/1705.07962}.

\bibitem[Si et~al.(2025)Si, Zhang, Li, Yang, Liu, and Yang]{si2025design2code}
Chenglei Si, Yanzhe Zhang, Ryan Li, Zhengyuan Yang, Ruibo Liu, and Diyi Yang.
\newblock Design2code: Benchmarking multimodal code generation for automated
  front-end engineering.
\newblock In \emph{Proceedings of the 2025 Conference of the Nations of the
  Americas Chapter of the Association for Computational Linguistics: Human
  Language Technologies (Volume 1: Long Papers)}, pages 3956--3974. Association
  for Computational Linguistics, 2025.
\newblock \doi{10.18653/v1/2025.naacl-long.199}.
\newblock URL \url{https://aclanthology.org/2025.naacl-long.199/}.

\bibitem[Yun et~al.(2024)Yun, Lin, Thushara, Bhat, Wang, Jiang, Deng, Wang,
  Tao, Li, Li, Nakov, Baldwin, Liu, Xing, Liang, and Shen]{yun2024web2code}
Sukmin Yun, Haokun Lin, Rusiru Thushara, Mohammad~Qazim Bhat, Yongxin Wang,
  Zutao Jiang, Mingkai Deng, Jinhong Wang, Tianhua Tao, Junbo Li, Haonan Li,
  Preslav Nakov, Timothy Baldwin, Zhengzhong Liu, Eric~P. Xing, Xiaodan Liang,
  and Zhiqiang Shen.
\newblock Web2code: A large-scale webpage-to-code dataset and evaluation
  framework for multimodal llms.
\newblock In \emph{Advances in Neural Information Processing Systems},
  volume~37, pages 112134--112157. Curran Associates, Inc., 2024.
\newblock \doi{10.52202/079017-3560}.
\newblock URL
  \url{https://proceedings.neurips.cc/paper_files/paper/2024/hash/cb66be286795d71f89367d596bf78ea7-Abstract-Datasets_and_Benchmarks_Track.html}.

\bibitem[Li et~al.(2020)Li, Luk{\'a}{\v c}, Gharbi, and
  Ragan-Kelley]{li2020diffvg}
Tzu-Mao Li, Michal Luk{\'a}{\v c}, Micha{\"e}l Gharbi, and Jonathan
  Ragan-Kelley.
\newblock Differentiable vector graphics rasterization for editing and
  learning.
\newblock \emph{ACM Transactions on Graphics}, 39\penalty0 (6):\penalty0 1--15,
  2020.
\newblock \doi{10.1145/3414685.3417871}.
\newblock URL \url{https://doi.org/10.1145/3414685.3417871}.

\bibitem[Carlier et~al.(2020)Carlier, Danelljan, Alahi, and
  Timofte]{carlier2020deepsvg}
Alexandre Carlier, Martin Danelljan, Alexandre Alahi, and Radu Timofte.
\newblock Deepsvg: A hierarchical generative network for vector graphics
  animation.
\newblock In \emph{Advances in Neural Information Processing Systems},
  volume~33, 2020.
\newblock URL
  \url{https://proceedings.neurips.cc/paper/2020/hash/bcf9d6bd14a2095866ce8c950b702341-Abstract.html}.

\bibitem[Ma et~al.(2022)Ma, Zhou, Xu, Sun, Filev, Orlov, Fu, and
  Shi]{ma2022live}
Xu~Ma, Yuqian Zhou, Xingqian Xu, Bin Sun, Valerii Filev, Nikita Orlov, Yun Fu,
  and Humphrey Shi.
\newblock Towards layer-wise image vectorization.
\newblock In \emph{Proceedings of the IEEE/CVF Conference on Computer Vision
  and Pattern Recognition}, pages 16314--16323, 2022.
\newblock \doi{10.1109/CVPR52688.2022.01583}.
\newblock URL
  \url{https://openaccess.thecvf.com/content/CVPR2022/html/Ma_Towards_Layer-Wise_Image_Vectorization_CVPR_2022_paper.html}.

\bibitem[Hu et~al.(2024)Hu, Yi, Qian, Zhang, Rosin, and Lai]{hu2024supersvg}
Teng Hu, Ran Yi, Baihong Qian, Jiangning Zhang, Paul~L. Rosin, and Yu-Kun Lai.
\newblock Supersvg: Superpixel-based scalable vector graphics synthesis.
\newblock In \emph{Proceedings of the IEEE/CVF Conference on Computer Vision
  and Pattern Recognition}, pages 24892--24901, 2024.
\newblock \doi{10.1109/CVPR52733.2024.02351}.
\newblock URL
  \url{https://openaccess.thecvf.com/content/CVPR2024/html/Hu_SuperSVG_Superpixel-based_Scalable_Vector_Graphics_Synthesis_CVPR_2024_paper.html}.

\bibitem[Yang et~al.(2025{\natexlab{b}})Yang, Cheng, Chen, Zeng, Yin, Zhang,
  Wang, Yu, Ma, and Jiang]{yang2025omnisvg}
Yiying Yang, Wei Cheng, Sijin Chen, Xianfang Zeng, Fukun Yin, Jiaxu Zhang, Liao
  Wang, Gang Yu, Xingjun Ma, and Yu-Gang Jiang.
\newblock Omnisvg: A unified scalable vector graphics generation model.
\newblock In \emph{Advances in Neural Information Processing Systems},
  2025{\natexlab{b}}.
\newblock URL \url{https://arxiv.org/abs/2504.06263}.

\bibitem[Nishina and Matsui(2024)]{nishina2024svgeditbench}
Kunato Nishina and Yusuke Matsui.
\newblock Svgeditbench: A benchmark dataset for quantitative assessment of
  {LLM}'s {SVG} editing capabilities.
\newblock In \emph{Proceedings of the IEEE/CVF Conference on Computer Vision
  and Pattern Recognition Workshops}, 2024.
\newblock URL \url{https://arxiv.org/abs/2404.13710}.

\bibitem[Hazimeh et~al.(2026)Hazimeh, Wang, Collier, Baechler, Kokiopoulou, and
  Frossard]{hazimeh2025slider}
Adam Hazimeh, Ke~Wang, Mark Collier, Gilles Baechler, Efi Kokiopoulou, and
  Pascal Frossard.
\newblock Semantic document derendering: {SVG} reconstruction via
  vision-language modeling.
\newblock \emph{Proceedings of the AAAI Conference on Artificial Intelligence},
  40\penalty0 (6):\penalty0 4636--4644, 2026.
\newblock \doi{10.1609/aaai.v40i6.42464}.
\newblock URL \url{https://ojs.aaai.org/index.php/AAAI/article/view/42464}.

\bibitem[Gonzalez(2026)]{gonzalez2026images2slides}
Leonardo Gonzalez.
\newblock From dead pixels to editable slides: Infographic reconstruction into
  native {Google Slides} via vision-language region understanding.
\newblock In \emph{Companion Proceedings of the ACM Web Conference 2026}, 2026.
\newblock URL \url{https://arxiv.org/abs/2602.07645}.

\bibitem[Hu et~al.(2026)Hu, Xue, Liang, Qi, Li, Wang, Xu, and
  Yu]{hu2026amodalsvg}
Juncheng Hu, Ziteng Xue, Guotao Liang, Anran Qi, Buyu Li, Sheng Wang, Dong Xu,
  and Qian Yu.
\newblock Amodalsvg: Amodal image vectorization via semantic layer peeling,
  2026.
\newblock URL \url{https://arxiv.org/abs/2604.10940}.

\bibitem[Jiang et~al.(2026)Jiang, Franke, Adesso, Haas, and
  Zhang]{jiang2026anchorflow}
Mengnan Jiang, Christian Franke, Michele~Franco Adesso, Antonio Haas, and
  Grace~Li Zhang.
\newblock Anchorflow: Editable {SVG} reconstruction via sparse anchor point
  fields, 2026.
\newblock URL \url{https://arxiv.org/abs/2605.19551}.

\bibitem[Su et~al.(2026)Su, Dong, Tang, Tang, Zhai, Lin, Chen, Gai, Luo, Wang,
  and Chu]{su2026vcgbench}
Xiaoyan Su, Peijie Dong, Zhenheng Tang, Song Tang, Yuyao Zhai, Kaitao Lin,
  Liang Chen, Yuhang Gai, Yuyu Luo, Qiang Wang, and Xiaowen Chu.
\newblock Vcg-bench: Towards a unified visual-centric benchmark for structured
  generation and editing, 2026.
\newblock URL \url{https://arxiv.org/abs/2605.15677}.

\bibitem[Deganutti et~al.(2026)Deganutti, Hirsch, Zhu, Seol, and
  Mehta]{deganutti2026gdb}
Adrienne Deganutti, Elad Hirsch, Haonan Zhu, Jaejung Seol, and Purvanshi Mehta.
\newblock Graphic-design-bench: A comprehensive benchmark for evaluating {AI}
  on graphic design tasks, 2026.
\newblock URL \url{https://arxiv.org/abs/2604.04192}.

\bibitem[Jeong et~al.(2026)Jeong, Byun, Son, Kim, and Kim]{jeong2025canvas}
Daeheon Jeong, Seoyeon Byun, Kihoon Son, Dae~Hyun Kim, and Juho Kim.
\newblock Canvas: A benchmark for vision-language models on tool-based user
  interface design.
\newblock \emph{Proceedings of the AAAI Conference on Artificial Intelligence},
  40\penalty0 (26):\penalty0 22182--22190, 2026.
\newblock \doi{10.1609/aaai.v40i26.39374}.
\newblock URL \url{https://ojs.aaai.org/index.php/AAAI/article/view/39374}.

\bibitem[Carion et~al.(2025)Carion, Gustafson, Hu, Debnath, Hu, Suris, Ryali,
  Alwala, Khedr, Huang, et~al.]{carion2025sam3}
Nicolas Carion, Laura Gustafson, Yuan-Ting Hu, Shoubhik Debnath, Ronghang Hu,
  Didac Suris, Chaitanya Ryali, Kalyan~Vasudev Alwala, Haitham Khedr, Andrew
  Huang, et~al.
\newblock {SAM 3}: Segment anything with concepts, 2025.
\newblock URL \url{https://arxiv.org/abs/2511.16719}.

\bibitem[Cui et~al.(2025)Cui, Sun, Lin, Gao, Zhang, Liu, Wang, Zhang, Zhou,
  Liu, et~al.]{cui2025paddleocr}
Cheng Cui, Ting Sun, Manhui Lin, Tingquan Gao, Yubo Zhang, Jiaxuan Liu, Xueqing
  Wang, Zelun Zhang, Changda Zhou, Hongen Liu, et~al.
\newblock {PaddleOCR 3.0} technical report, 2025.
\newblock URL \url{https://arxiv.org/abs/2507.05595}.

\end{thebibliography}

\appendix

\clearpage
\begin{center}
{\RRtitlefont\LARGE\color{RRInk}DrawAI: Supplementary Material}
\end{center}
\section{Complete Evaluation Criteria}
\label{sec:complete-evaluation-criteria}

This section specifies the complete criterion-level implementation of DrawAI-Bench. Let $I,R\in[0,1]^{H\times W\times 3}$ denote the source image and candidate rendering after both are aligned to the annotated canvas. Let $G_t$ and $\widehat G_t$ denote the ground-truth and candidate objects of asset type $t$, $B_g$ the annotated box of element $g$, and $M$ a greedy one-to-one match set. All formulas in Table~\ref{tab:complete-evaluation-criteria} produce raw criterion scores. Before aggregation, valid scores are clipped to $[0,1]$, with metrics marked $\downarrow$ first converted to $1-r$.

For global structural similarity, we use
\begin{equation}
\operatorname{SSIM}(I,R)=
\frac{(2\mu_I\mu_R+C_1)(2\sigma_{IR}+C_2)}
     {(\mu_I^2+\mu_R^2+C_1)(\sigma_I^2+\sigma_R^2+C_2)}.
\label{eq:appendix-ssim}
\end{equation}
For edge alignment, Canny edges $E_I$ and $E_R$ are compared with a one-pixel dilation $\delta(\cdot)$, giving $P_e=|E_R\cap\delta(E_I)|/|E_R|$ and $R_e=|E_I\cap\delta(E_R)|/|E_I|$. Text content is matched one-to-one after Unicode, case, and whitespace normalization. The default thresholds are 0.8 text similarity, 0.5 text-box IoU, 0.5 image IoU, and 0.02 region MSE for the localized-image fallback. Split SVG text lines may be grouped before text-box IoU is computed, and unmatched ground-truth image boxes contribute zero to image mIoU.

Every rubric receives only the source image, candidate rendering, ground-truth annotations, and candidate SVG. Visual judgments use the rendering, whereas editability judgments use SVG structure. For criterion $m$, the evaluator maps its answer to
\begin{equation}
q_m(e)=
\begin{cases}
1, & \text{yes},\\
0.5, & \text{partial},\\
0, & \text{no},\\
\bot, & \text{N/A},
\end{cases}
\qquad
s_m=\frac{1}{|G_t^+|}\sum_{e\in G_t^+}q_m(e),
\label{eq:appendix-rubric-score}
\end{equation}
where $G_t^+=\{e\in G_t:q_m(e)\neq\bot\}$. Global rubric criteria replace $e$ with the complete page and produce one judgment. The questions in the table retain the operational meaning of the full evaluator prompts while omitting file paths, output schemas, and execution instructions.

\begingroup
\setlength{\tabcolsep}{2.5pt}
\renewcommand{\arraystretch}{1.05}
\begin{longtable}{>{\centering\arraybackslash}p{0.035\textwidth}>{\raggedright\arraybackslash}p{0.15\textwidth}>{\centering\arraybackslash}p{0.075\textwidth}>{\raggedright\arraybackslash}p{0.25\textwidth}>{\raggedright\arraybackslash}p{0.42\textwidth}}
\caption{Complete DrawAI-Bench criteria, raw computations, and simplified rubric questions. Rule values are raw; $\downarrow$ metrics use $1-r$ before aggregation.}
\label{tab:complete-evaluation-criteria}\\
\toprule
\# & Criterion & Mode & Raw criterion score & Simplified QA question \\
\midrule
\endfirsthead
\multicolumn{5}{l}{\tablename~\thetable\ (continued)}\\[2pt]
\toprule
\# & Criterion & Mode & Raw criterion score & Simplified QA question \\
\midrule
\endhead
\midrule
\multicolumn{5}{r}{\scriptsize Continued on next page}\\
\endfoot
\bottomrule
\endlastfoot
\rowcolor{DrawAIHeatBlue!16}
\multicolumn{5}{l}{\strut\textbf{Fidelity: Global Visual Fidelity}}\\*
\rowcolor{DrawAIHeatBlue!3}
1 & Structural match & Rule $\uparrow$ & Eq.~\ref{eq:appendix-ssim} & Deterministic; no VLM question. \\
\rowcolor{DrawAIHeatBlue!3}
2 & Edge alignment & Rule $\uparrow$ & \(s_m=2P_eR_e/(P_e+R_e)\), with tolerance-aware edge precision and recall & Deterministic; no VLM question. \\
\rowcolor{DrawAIHeatBlue!3}
3 & Pixel error & Rule $\downarrow$ & \(s_m=(3HW)^{-1}\lVert I-R\rVert_2^2\) & Deterministic; no VLM question. \\
\rowcolor{DrawAIHeatBlue!3}
4 & Overall readability & Rubric & \(s_m=q_m(\mathrm{page})\) & Compare the complete reference and rendered candidate. Is the whole result readable and coherent, with important content legible? \\
\rowcolor{DrawAIHeatBlue!3}
5 & Layout skeleton & Rubric & \(s_m=q_m(\mathrm{page})\) & Are the major regions, flow, alignment, proportions, and overall composition skeleton preserved? \\
\rowcolor{DrawAIHeatBlue!3}
6 & Visual hierarchy & Rubric & \(s_m=q_m(\mathrm{page})\) & Are emphasis, contrast, visual hierarchy, and the absence of distracting artifacts preserved? \\
\rowcolor{DrawAIHeatBlue!16}
\multicolumn{5}{l}{\strut\textbf{Fidelity: Text Fidelity}}\\*
\rowcolor{DrawAIHeatBlue!3}
7 & Text content & Rule $\uparrow$ & \(P_t=|M_t|/|\widehat G_t|\), \(R_t=|M_t|/|G_t|\), \(s_m=2P_tR_t/(P_t+R_t)\) & Deterministic; no VLM question. \\
\rowcolor{DrawAIHeatBlue!3}
8 & Position lines & Rubric & \(s_m=|G_t^+|^{-1}\sum_{e\in G_t^+}q_m(e)\) & For this GT text element, is the text in the correct region with equivalent line grouping and line breaks? \\
\rowcolor{DrawAIHeatBlue!3}
9 & Alignment spacing & Rubric & \(s_m=|G_t^+|^{-1}\sum_{e\in G_t^+}q_m(e)\) & Is its alignment and spacing relative to its box and neighboring elements preserved? \\
\rowcolor{DrawAIHeatBlue!3}
10 & Overflow occlusion & Rubric & \(s_m=|G_t^+|^{-1}\sum_{e\in G_t^+}q_m(e)\) & Is it visible without overflow, clipping, collision, or occlusion? \\
\rowcolor{DrawAIHeatBlue!3}
11 & Color emphasis & Rubric & \(s_m=|G_t^+|^{-1}\sum_{e\in G_t^+}q_m(e)\) & Are text color and emphasis relationships visually faithful? \\
\rowcolor{DrawAIHeatBlue!3}
12 & Font appearance & Rubric & \(s_m=|G_t^+|^{-1}\sum_{e\in G_t^+}q_m(e)\) & Are approximate font-family appearance, size, weight, and styling faithful? \\
\rowcolor{DrawAIHeatBlue!3}
13 & Style consistency & Rubric & \(s_m=|G_t^+|^{-1}\sum_{e\in G_t^+}q_m(e)\) & Is its style consistent with other text elements of the same visual class? \\
\rowcolor{DrawAIHeatBlue!16}
\multicolumn{5}{l}{\strut\textbf{Fidelity: Image Fidelity}}\\*
\rowcolor{DrawAIHeatBlue!3}
14 & Image Regions & Rule $\downarrow$ & \(s_m=|G_{\rm img}|^{-1}\sum_g (3|B_g|)^{-1}\lVert I_{B_g}-R_{B_g}\rVert_2^2\) & Deterministic; no VLM question. \\
\rowcolor{DrawAIHeatBlue!3}
15 & Image Alignment & Rule $\uparrow$ & \(s_m=|G_{\rm img}|^{-1}\sum_g \operatorname{IoU}(B_g,\widehat B_{\pi(g)})\); unmatched \(=0\) & Deterministic; no VLM question. \\
\rowcolor{DrawAIHeatBlue!16}
\multicolumn{5}{l}{\strut\textbf{Fidelity: Formula Fidelity}}\\*
\rowcolor{DrawAIHeatBlue!3}
16 & Symbol retention & Rubric & \(s_m=|G_t^+|^{-1}\sum_{e\in G_t^+}q_m(e)\) & For this GT formula element, are its visible symbols, operators, labels, and key notation retained? \\
\rowcolor{DrawAIHeatBlue!3}
17 & Structure relation & Rubric & \(s_m=|G_t^+|^{-1}\sum_{e\in G_t^+}q_m(e)\) & Are superscripts, subscripts, fractions, grouping, baselines, and symbol relationships reconstructed coherently? \\
\rowcolor{DrawAIHeatBlue!3}
18 & Position readability & Rubric & \(s_m=|G_t^+|^{-1}\sum_{e\in G_t^+}q_m(e)\) & Is the formula in the correct region, legible, and free from clipping, overlap, or destructive spacing changes? \\
\rowcolor{DrawAIHeatBlue!16}
\multicolumn{5}{l}{\strut\textbf{Fidelity: Shape Fidelity}}\\*
\rowcolor{DrawAIHeatBlue!3}
19 & Existence coverage & Rubric & \(s_m=|G_t^+|^{-1}\sum_{e\in G_t^+}q_m(e)\) & Is the GT shape visibly reconstructed in the correct area? \\
\rowcolor{DrawAIHeatBlue!3}
20 & Visual attributes & Rubric & \(s_m=|G_t^+|^{-1}\sum_{e\in G_t^+}q_m(e)\) & Are fill, stroke, corner style, opacity, and overall appearance faithful? \\
\rowcolor{DrawAIHeatBlue!3}
21 & Neighbor relation & Rubric & \(s_m=|G_t^+|^{-1}\sum_{e\in G_t^+}q_m(e)\) & Are its spatial and containment relationships with neighboring elements preserved? \\
\rowcolor{DrawAIHeatBlue!16}
\multicolumn{5}{l}{\strut\textbf{Fidelity: Connector Fidelity}}\\*
\rowcolor{DrawAIHeatBlue!3}
22 & Path presence & Rubric & \(s_m=|G_t^+|^{-1}\sum_{e\in G_t^+}q_m(e)\) & Is the GT connector visibly present with the intended route? \\
\rowcolor{DrawAIHeatBlue!3}
23 & Endpoint direction & Rubric & \(s_m=|G_t^+|^{-1}\sum_{e\in G_t^+}q_m(e)\) & Do its endpoints, attachment regions, and direction or arrow orientation match? \\
\rowcolor{DrawAIHeatBlue!3}
24 & Line style noise & Rubric & \(s_m=|G_t^+|^{-1}\sum_{e\in G_t^+}q_m(e)\) & Are stroke style, dash, width, arrowhead, and routing faithful without extra line noise? Answer yes when the line is faithful and clean. \\
\rowcolor{DrawAIHeatBlue!16}
\multicolumn{5}{l}{\strut\textbf{Fidelity: Table Fidelity}}\\*
\rowcolor{DrawAIHeatBlue!3}
25 & Grid structure & Rubric & \(s_m=|G_t^+|^{-1}\sum_{e\in G_t^+}q_m(e)\) & For this GT table element, are the visible rows, columns, dividers, merged regions, and outer structure reconstructed? \\
\rowcolor{DrawAIHeatBlue!3}
26 & Content ownership & Rubric & \(s_m=|G_t^+|^{-1}\sum_{e\in G_t^+}q_m(e)\) & Do the visible labels and values belong to the correct cells or table regions without leaking into neighbors? \\
\rowcolor{DrawAIHeatBlue!3}
27 & Readability alignment & Rubric & \(s_m=|G_t^+|^{-1}\sum_{e\in G_t^+}q_m(e)\) & Is the table readable, with faithful cell alignment, spacing, ordering, and no material clipping or overlap? \\
\rowcolor{DrawAIHeatRed!16}
\multicolumn{5}{l}{\strut\textbf{Editability: Text Editability}}\\*
\rowcolor{DrawAIHeatRed!3}
28 & Text boxes & Rule $\uparrow$ & \(s_m=|M_{\rm box}|/|G_{\rm text}|\) & Deterministic; no VLM question. \\
\rowcolor{DrawAIHeatRed!16}
\multicolumn{5}{l}{\strut\textbf{Editability: Image Editability}}\\*
\rowcolor{DrawAIHeatRed!3}
29 & Image Sources & Rule $\uparrow$ & \(s_m=(|M_{\rm image}|+|M_{\rm MSE}|)/|G_{\rm img}|\) & Deterministic; no VLM question. \\
\rowcolor{DrawAIHeatRed!16}
\multicolumn{5}{l}{\strut\textbf{Editability: Shape Editability}}\\*
\rowcolor{DrawAIHeatRed!3}
30 & Object independence & Rubric & \(s_m=|G_t^+|^{-1}\sum_{e\in G_t^+}q_m(e)\) & Does \texttt{candidate.svg} provide an independently editable SVG object or coherent group for this shape, rather than baking it into unrelated content? \\
\rowcolor{DrawAIHeatRed!3}
31 & Box placement & Rubric & \(s_m=|G_t^+|^{-1}\sum_{e\in G_t^+}q_m(e)\) & Does the editable object occupy the correct GT bbox with reasonable placement and size? \\
\rowcolor{DrawAIHeatRed!3}
32 & Conflict diagnosis & Rubric & \(s_m=|G_t^+|^{-1}\sum_{e\in G_t^+}q_m(e)\) & Is the object cleanly separated without incorrect merges, splits, duplicate ownership, or object conflicts? Answer yes when there is no material conflict. \\
\rowcolor{DrawAIHeatRed!16}
\multicolumn{5}{l}{\strut\textbf{Editability: Connector Editability}}\\*
\rowcolor{DrawAIHeatRed!3}
33 & Independent path & Rubric & \(s_m=|G_t^+|^{-1}\sum_{e\in G_t^+}q_m(e)\) & Is it represented by an independent editable line, path, polyline, or coherent connector group in \texttt{candidate.svg}? \\
\rowcolor{DrawAIHeatRed!3}
34 & Endpoint control & Rubric & \(s_m=|G_t^+|^{-1}\sum_{e\in G_t^+}q_m(e)\) & Can its endpoints and route be edited coherently as connector geometry? \\
\rowcolor{DrawAIHeatRed!3}
35 & Mis merge check & Rubric & \(s_m=|G_t^+|^{-1}\sum_{e\in G_t^+}q_m(e)\) & Is it not incorrectly merged into a shape, text, raster image, or unrelated connector? Answer yes when it is correctly separate. \\
\rowcolor{DrawAIHeatRed!16}
\multicolumn{5}{l}{\strut\textbf{Editability: Formula Editability}}\\*
\rowcolor{DrawAIHeatRed!3}
36 & Editable representation & Rule $\uparrow$ & \(s_m=|M_{\rm native\mbox{-}formula}|/|G_{\rm formula}|\) & Deterministic; no VLM question. \\
\rowcolor{DrawAIHeatRed!16}
\multicolumn{5}{l}{\strut\textbf{Editability: Table Editability}}\\*
\rowcolor{DrawAIHeatRed!3}
37 & Table region independence & Rubric & \(s_m=|G_t^+|^{-1}\sum_{e\in G_t^+}q_m(e)\) & Does \texttt{candidate.svg} provide an independently locatable editable group, grid, line, rect, or text structure for this table region, rather than baking it into an unrelated full-page raster image? \\
\rowcolor{DrawAIHeatRed!3}
38 & Editable grid & Rubric & \(s_m=|G_t^+|^{-1}\sum_{e\in G_t^+}q_m(e)\) & Does \texttt{candidate.svg} provide independently editable grid geometry or a coherent editable table group? \\
\rowcolor{DrawAIHeatRed!3}
39 & Text ownership & Rubric & \(s_m=|G_t^+|^{-1}\sum_{e\in G_t^+}q_m(e)\) & Are table text objects independently editable and associated with the correct table region or cell? \\
\end{longtable}
\endgroup

\clearpage
\section{Simple Baseline Reconstruction Prompt}
\label{sec:reconstruction-prompt}

The Simple Baseline receives only the source raster and the shared image-to-editable-SVG prompt below. The prompt fixes the permitted operations, reconstruction procedure, Fidelity requirements, and required output files across model--harness settings.

\begin{drawaipromptbox}
\small
\drawaiprompttitle{Image-to-Editable-SVG Reconstruction Protocol}

\drawaipromptheading{Role}

Given a single source raster image, reconstruct it as a standalone, visually faithful, and structurally editable SVG document.

\drawaipromptheading{Objective}

Create an SVG whose rendered appearance closely matches the source image while preserving semantic structure and editability wherever practical. The task must be completed by inspecting the supplied image and writing and executing local code within the run directory. Do not merely describe a solution, return a declarative scene specification, or stop at an intermediate representation.

\drawaipromptheading{Permitted Operations}

\begin{itemize}
\tightlist
\item
  Use the selected agent/model as the only model for visual understanding, reasoning, and implementation.
\item
  Write local scripts and intermediate files required to construct, render, and validate the SVG.
\item
  Use native SVG elements, including text, groups, paths, rectangles, circles, ellipses, polygons, polylines, lines, gradients, clipping paths, masks, and filters when appropriate.
\item
  Use deterministic crops from the source image for complex photos, illustrations, textures, screenshots, icons, logos, or other regions that cannot be faithfully reconstructed as vector objects.
\end{itemize}

\drawaiprompttitle{Reconstruction Procedure}

\begin{enumerate}
\def\labelenumi{\arabic{enumi}.}
\tightlist
\item
  Inspect the complete source image and infer its canvas dimensions, visual hierarchy, major regions, text, spatial relationships, colors, and reusable graphical structures.
\item
  Decompose the slide into semantic components. Distinguish content that should be represented as editable SVG text and vector primitives from complex raster content that should be preserved through localized source crops.
\item
  Set the SVG \texttt{viewBox} to match the source dimensions or an exactly proportional coordinate system, then reconstruct the slide in visual stacking order: background, large regions, structural elements, text, diagrams, and localized raster assets.
\item
  Preserve text as SVG \texttt{\textless{}text\textgreater{}} and \texttt{\textless{}tspan\textgreater{}} elements when it is readable. Preserve grouping, alignment, line breaks, emphasis, and approximate font metrics.
\item
  When local deterministic SVG rendering is available, render \path{output/result.svg} to \path{output/preview.png}, compare it with the source image, and revise discrepancies in element presence, layout, text placement, scale, color, style, stacking order, and clarity.
\item
  Generate the final SVG only after checking that it is valid, standalone, renderable, and visually faithful.
\end{enumerate}

\drawaipromptheading{Fidelity Criteria}

\begin{itemize}
\tightlist
\item
  Preserve the source slide\textquotesingle s aspect ratio, composition, alignment, spacing, typography hierarchy, colors, borders, and relative element sizes as closely as practical.
\item
  Use semantic SVG primitives rather than converting all content into undifferentiated paths.
\item
  Represent simple shapes, connectors, arrows, tables, charts, diagrams, and icons as native editable SVG elements whenever possible.
\item
  Prefer a localized deterministic crop over an inaccurate vector simplification for complex raster artwork.
\item
  Do not use the whole source image, or a near-full-page crop, as an SVG \texttt{\textless{}image\textgreater{}} background.
\item
  Avoid omitted elements, unintended overlaps, clipped text, malformed paths, broken asset references, and generic placeholder graphics.
\end{itemize}

\drawaipromptheading{Output Contract}

Produce the following files:

\begin{itemize}
\tightlist
\item
  \texttt{output/result.svg}: the final standalone SVG reconstruction.
\item
  \texttt{output/notes.md}: a concise record of the reconstruction strategy, native SVG elements, source crops, and known limitations.
\item
  \texttt{output/preview.png}: a deterministic local rendering of the final SVG when a renderer is available.
\item
  \texttt{output/agent-result.json}: the completion sentinel, written last.
\end{itemize}

\end{drawaipromptbox}
\refstepcounter{figure}
\par\smallskip\noindent
{\small\RRsans\bfseries Figure~\thefigure.}\hspace{0.35em}
{\small Markdown prompt template used by the Simple Baseline for image-to-editable-SVG reconstruction.}
\label{fig:reconstruction-prompt}

\end{document}